%% file: acl_latex.tex
\pdfoutput=1

\documentclass[11pt]{article}

\usepackage{acl}

\usepackage{times}
\usepackage{latexsym}

\usepackage[T1]{fontenc}

\usepackage[utf8]{inputenc}

\usepackage{microtype}

\usepackage{inconsolata}

\usepackage{graphicx}

\usepackage{inconsolata}
\usepackage{graphicx}

\usepackage{enumitem}
\usepackage{multirow}
\usepackage{booktabs}
\usepackage{algorithm}
\usepackage{algorithmic}
\usepackage{tikz}
\usepackage{pgfplots}
\usepackage{graphicx}
\usepackage{subcaption}
\usepackage{pgf-pie}
\usepackage{enumitem}
\usepackage{pifont}
\usepackage{amsmath}
\usepackage{supertabular}
\usepackage{booktabs}
\numberwithin{equation}{section}
\usepackage{dsfont}
\usepgfplotslibrary{polar}

\pgfplotsset{compat=1.18}
\usepackage{bm}
\usepackage{xspace}
\usepackage{tcolorbox}
\usepackage{bbding}
\usepackage{wasysym}
\usepackage{amssymb}
\usepackage{fontawesome5}
\usepackage{arydshln}  
\usepackage{dashrule}

\newcommand{\ourmethod}{\textsc{Ours}\xspace}
\newcommand{\ourdataset}{\textsc{MultiTAT}\xspace}
\definecolor{cpurple}{rgb}{0.675, 0.573, 0.922}
\definecolor{cblue}{rgb}{0.310, 0.757, 0.910}
\definecolor{cgreen}{rgb}{0.310, 0.563, 0.214}
\definecolor{corange}{rgb}{1, 0.808, 0.329}
\definecolor{cred}{rgb}{0.8, 0.3, 0.3}
\definecolor{data_blue}{rgb}{0.809, 0.883, 0.949}
\definecolor{data_blue_light}{rgb}{0.930, 0.965, 1}
\definecolor{data_blue_dark}{rgb}{0.027, 0.215, 0.387}
\definecolor{annotator_pink}{rgb}{0.914, 0.816, 0.859}
\definecolor{gray}{rgb}{0.715, 0.715, 0.715}
\definecolor{gray_light}{rgb}{0.949, 0.949, 0.949}
\definecolor{reasoner_green}{rgb}{0.848, 0.914, 0.824}
\newcommand{\datatext}[1]{{\color{data_blue_dark}#1}}
\newcommand{\textgray}[1]{{\color{gray}#1}}

\newtcolorbox{myexample}[2][]{
  colback=data_blue!40,
  colframe=data_blue,         
  coltitle=black,
  title=\textbf{#2},
  fonttitle=\bfseries,
  #1,
}

\title{\ourdataset: Benchmarking Multilingual Table-and-Text\\Question Answering}

\author{
    Xuanliang Zhang\footnotemark[2], Dingzirui Wang\footnotemark[2], Keyan Xu, Qingfu Zhu, Wanxiang Che\\ 
    \texttt{\{xuanliangzhang, dzrwang, kyxu, qfzhu, car\}@ir.hit.edu.cn}\\
    Harbin Institute of Technology 
}

\begin{document}
    \maketitle

\renewcommand{\thefootnote}{\fnsymbol{footnote}}
\footnotetext[2]{Equal contribution.}
\renewcommand{\thefootnote}{\arabic{footnote}}
    \begin{abstract}
    Question answering on the hybrid context of tables and text (TATQA) is a critical task, with broad applications in data-intensive domains.
    However, existing TATQA datasets are limited to English, leading to several drawbacks:
    (i)~They overlook the challenges of multilingual TAT-QA and cannot assess model performance in the multilingual setting.
    (ii)~They do not reflect real-world scenarios where tables and texts frequently appear in non-English languages.
    To address the limitations, we propose the first multilingual TATQA dataset (\ourdataset).
    Specifically, we sample data from $3$ mainstream TATQA datasets and translate it into $10$ diverse languages.
    To align the model TATQA capabilities in English with other languages, we develop a baseline, \ourmethod.
    Experimental results reveal that the performance on non-English data in \ourdataset drops by an average of $19.4\%$ compared to English, proving the necessity of \ourdataset.
    We further analyze the reasons for this performance gap.
    Furthermore, \ourmethod outperforms other baselines by an average of $3.3$, demonstrating its effectiveness\footnote{Our data is available at \href{https://github.com/zhxlia/MULTITAT}{github.com/zhxlia/MULTITAT}}.


    \end{abstract}

    \section{Introduction}
        \input{tex/1.introduction}

    \section{\ourdataset}
        \label{sec:dataset}
        \input{tex/2.dataset}

    \section{\ourmethod}
        \label{sec:methodology}
        \input{tex/3.methodology}
    
    \section{Experiments}
        \label{sec:experiments}
        \input{tex/4.experiment}
    
    \section{Related Works}
        \label{sec:related}
        \input{tex/5.related}
    
    \section{Conclusion}
        To address the limitations of the existing QA datasets on the hybrid context of tabular and text data (TATQA), which focus exclusively on a single language, we introduce the first multilingual TAT-QA dataset \ourdataset. 
        Specifically, we sample data from mainstream TAT-QA datasets, including HybridQA, TAT-QA, and SciTAT, and translate it into $10$ diverse languages.
        To enhance the TATQA performance in non-English languages, we propose a baseline (\ourmethod). 
        \ourmethod links the relevant information from the hybrid context and reasons in English.
        We conduct a series of baseline experiments and observe a $19.4\%$ performance drop for non-English languages compared to English. 
        Error analysis revealed that this decline is primarily due to the increased difficulty in linking relevant information in non-English texts and the reduced ability to apply formulas and follow instructions of models.
        Furthermore, \ourmethod achieves an average improvement of $3.3$ over other baselines, demonstrating its effectiveness. 
        Analysis of experimental results suggests that the performance of TATQA across languages is influenced not only by high-resource versus low-resource languages but also by the inherent characteristics of the model itself.

    \clearpage
    
    \section*{Limitations}
        (\emph{i})~\ourdataset only includes single-turn dialogues, leaving multilingual multi-turn dialogues for future work.
        (\emph{ii})~\ourdataset covers only $11$ languages. Future versions should include more languages.
        
    \section*{Ethics Statement}
            
    All datasets and models used in this paper are publicly available, and our utilization of them strictly complies with their respective licenses and terms of use. 
    Additionally, we confirm that the compensation provided to annotators is significantly higher than the local minimum wage.
    
    \bibliography{custom}
    
    \clearpage
    \appendix
    \label{sec:appendix}
    \input{tex/6.appendix}
    
\end{document}

%% file: tex/1.introduction.tex
Question answering over the hybrid context of tabular and textual data (TATQA) is an important task \cite{chen-etal-2020-hybridqa}, which is widely used in data-intensive fields, such as finance and science, gaining increasing attention \cite{chen-etal-2021-finqa,auer2023sciqa}. 
Enhancing the TATQA capabilities of models can significantly aid in extracting useful information from hybrid data.
The heterogeneous evidence brings challenges to the TATQA task since it requires the model to link the relevant information in the table or text according to the entities in the question \cite{feng-etal-2022-multi-hop,lei-etal-2022-Graph-based-Encoder,wang2022hybridqa-survey}.

\begin{figure}[t]
    \centering
    \includegraphics[width=1.\linewidth]{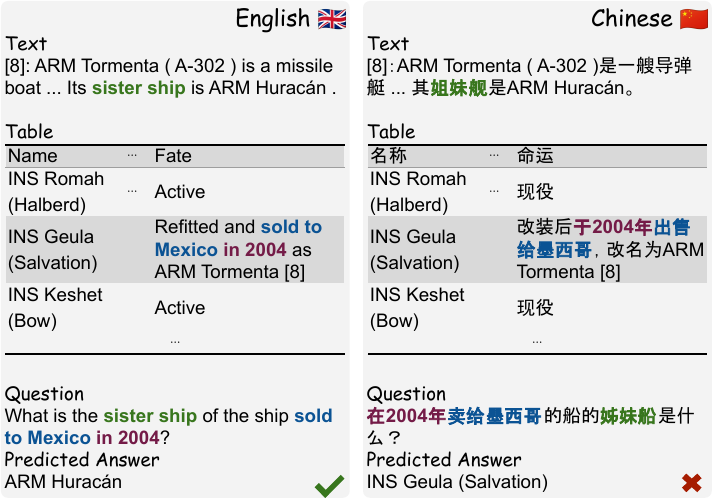}
    \vspace{-0.5em}
    \caption{
    Comparison of the English and Chinese examples in \ourdataset.
    Entities with the same color annotation represent corresponding entity information.
    In Chinese, the richness of lexical expressions makes it more challenging for the model to link relevant information, leading to the incorrect predicted answer.
    }
    \label{fig:intro}
\end{figure}

To evaluate the model capabilities on the TATQA task, several datasets are proposed \cite{li2021tsqa,chen-etal-2021-finqa,zhao-etal-2024-docmath}. 
For example, HybridQA~\cite{chen-etal-2020-hybridqa}, TAT-QA~\cite{zhu-etal-2021-tat}, and SciTAT~\cite{zhang2024scitat} respectively construct English TATQA datasets in the domains of Wikipedia, finance, and science. 
However, these datasets focus solely on English, having the following shortcomings:
(\emph{i})~They \textbf{\textit{cannot adequately assess the TATQA performance in the multilingual setting, overlooking the challenges of multilingual TATQA}}.
As shown in Figure~\ref{fig:intro}, the complex lexical expressions of different languages pose challenges for models to link information across hybrid contexts \cite{MultiSpider}.
(\emph{ii})~They \textbf{\textit{create a gap with real-world scenarios}}, as domains such as finance and science contain substantial amounts of non-English tables and text \cite{hamotskyi-etal-2024-fincorpus,angulo2021non-English-scientific,bhagavatula-etal-2012-language-wiki}.
To address the limitations, 
we propose the first multilingual TATQA benchmark, comprising parallel data in $11$ diverse languages.

First, we introduce the multilingual TATQA dataset (\ourdataset).
To ensure the high quality of \ourdataset, we sample data from three mainstream English TATQA datasets and employ a combination of machine translation and manual revision to translate them into $10$ languages.
In total, \ourdataset consists of $250$ questions from $233$ hybrid contexts, covering three domains: Wikipedia, finance, and science. 

To enhance the performance on \ourdataset of non-English languages, we propose a baseline to bridge the performance gap between English and non-English on TATQA (\ourmethod). 
To align the model TATQA capabilities in English with other languages, especially low-resource languages, \ourmethod is divided into two modules: linking non-English information and reasoning in English. 
Specifically, \ourmethod first identifies relevant information from tables and text according to the entities in the question through linking and then uses this information to perform reasoning in English by generating programs.

We evaluate the performance of \ourmethod, compared with a series of baselines on \ourdataset. 
Experimental results indicate that the performance of non-English languages drops by an average of $19.4\%$ compared to English on all baselines, highlighting the necessity of \ourdataset. 
\ourmethod outperforms other baselines by an average of $3.3$, demonstrating the effectiveness. 
Analysis experiments reveal that the TATQA capabilities across languages are not only influenced by resource availability but also by their specific linguistic characteristics. 
Error analysis shows that the performance decline in non-English TATQA is primarily due to the reduced ability to link relevant information, apply formulas, and follow instructions.

Our contributions are as follows:
\begin{enumerate}
    \item To the best of our knowledge, we introduce the first multilingual TATQA dataset \ourdataset, which includes $11$ diverse languages.
    \item We propose \ourmethod, a baseline to align the model TATQA capabilities in English to non-English languages.
    \item We conduct a series of experiments, supported by empirical results and error analysis, to demonstrate the challenges of \ourdataset and provide insights for future improvements.
\end{enumerate}

%% file: tex/2.dataset.tex
\begin{figure*}
    \centering
    \includegraphics[width=.85\linewidth]{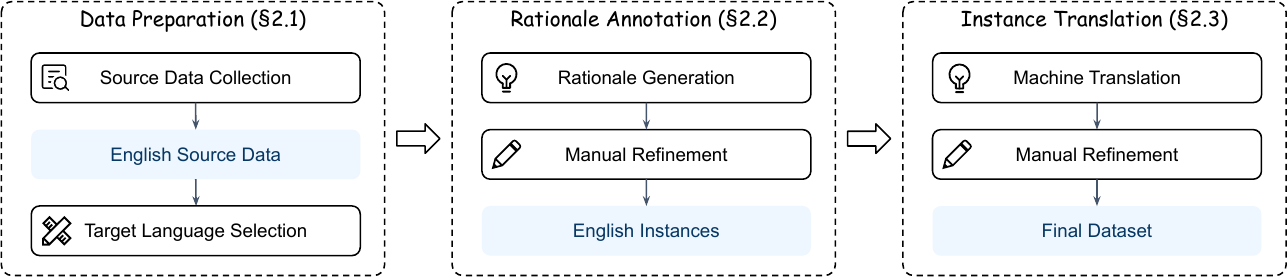}
    \caption{
    The process of constructing \ourdataset.
    The \colorbox{data_blue_light}{\datatext{blue}} boxes represent the data, and the white solid boxes represent the construction steps.
    }
    \label{fig:framework}
\end{figure*}

The input of \ourdataset consists of a question, the hybrid context including the table and text, and the output is the answer to the question. 
Additionally, we annotate the rationale, which is the reasoning process of answering the question in \ourdataset. 
We refer to each question, along with its corresponding table, text, rationale, and answer, as an instance. 
For each instance, we annotate $11$ diverse languages.
We first describe the construction process of \ourdataset, which combines automatic generation with manual error correction, following previous works \cite{peng-etal-2024-humanevalxl,singh-etal-2024-indicgenbench,MultiSpider}, as shown in Figure~\ref{fig:framework}.

\begin{table*}[t]
    \centering
    \small
    \input{tab/answer_statistics}
    \caption{
    The distribution of English data, including answer types and answer sources in \ourdataset, sourced from three mainstream datasets.
    The listed answer types are the all answer types corresponding to each dataset.
    }
    \label{tab:answer_statistics}
\end{table*}

\subsection{Data Preparation}
We first collect English data from existing datasets and select languages to translate them.

\subsubsection{Source Data Collection}
We select HybridQA~\cite{chen-etal-2020-hybridqa}, TAT-QA~\cite{zhu-etal-2021-tat}, and SciTAT~\cite{zhang2024scitat} datasets from the Wikipedia, finance, and science domains as our data sources, as these three domains are the primary areas where TATQA tasks are currently distributed (see Table~\ref{tab:comparison_tat}). 
To ensure an even distribution of different answer types and answer sources in \ourdataset, we sample a total of $250$ instances from the three datasets according to the proportions shown in Table~\ref{tab:answer_statistics}. 
Among them, only $50$ instances are sampled from HybridQA due to its relatively limited answer sources and types. 

\subsubsection{Target Language Selection}
For \ourdataset, we select $11$ languages, covering $8$ language families: Bengali (bn), Chinese (zh), English (en), French (fr), German (de), Japanese (ja), Russian (ru), Spanish (es), Swahili (sw), Telugu (te), and Thai (th), following the previous benchmark \cite{shi2023MGSM}. 
Additionally, we preserve the Arabic numerals from the original datasets across all languages to facilitate evaluation \cite{shi2023MGSM}.

\subsection{Rationale Annotation}
\label{subsec:Rationale Annotation}
We first demonstrate how to annotate English rationales by employing the large language model (LLM) in combination with manual refinement. 
We use \texttt{gpt-4o}~\cite{openai2024gpt4technicalreport} to complete \textbf{rationale generation} due to its strong reasoning and instruction-following capabilities. 
Specifically, we input the question, relevant tables and texts, and the answer into the LLM, prompting the LLM to generate the corresponding rationale.
Since LLMs cannot guarantee the accuracy of reasoning, we employ \textbf{manual refinement}. 
The annotators are instructed to evaluate the accuracy of the generated rationale and make corrections where necessary.

\subsection{Instance Translation}
\label{subsec:Instance Translation}
In this section, we describe the process of combining the LLM with human annotations to translate English instances into $10$ languages. 
For \textbf{machine translation}, we select \texttt{gpt-4o} because of its strong translation capabilities \cite{yan2024gpt4vshumantranslators,hu-etal-2024-gentranslate}. 
Specifically, we input each instance into the LLM, with prompts to translate it into the target languages, respectively. 
To assess the accuracy of the translations, we use \texttt{gpt-4o} to translate the target language instances back into English, and calculate the F1 score between the back-translated version and the original English instance following previous works \cite{peng-etal-2024-humanevalxl}. 
For instances with an F1 score below $0.6$, we prompt annotators to complete \textbf{manual refinement} by using Google Translation for a new translation.

\subsection{Quality Control}
To ensure the quality of \ourdataset, we implement rigorous quality control strategies.
The annotators we hire hold graduate-level degrees, are proficient in English and are compensated with $\$1$ per data instance. 
We first train the annotators to familiarize them with the annotation requirements and the use of the annotation tool (see Appendix~\ref{subsec:Annotator Training Process}). 
Then, they try to annotate $20$ instances, and we review their annotations, providing feedback and suggestions for revisions.



\subsection{Data Analysis}
We show the data distribution of \ourdataset in Table~\ref{tab:answer_statistics}.
The $250$ questions in \ourdataset involve $233$ hybrid contexts, each of which includes $1$ table and an average of $5.3$ paragraphs. 
Each table has an average of $10.2$ rows and $4.7$ columns.

\begin{figure*}[t]
    \centering
    \includegraphics[width=.95\linewidth]{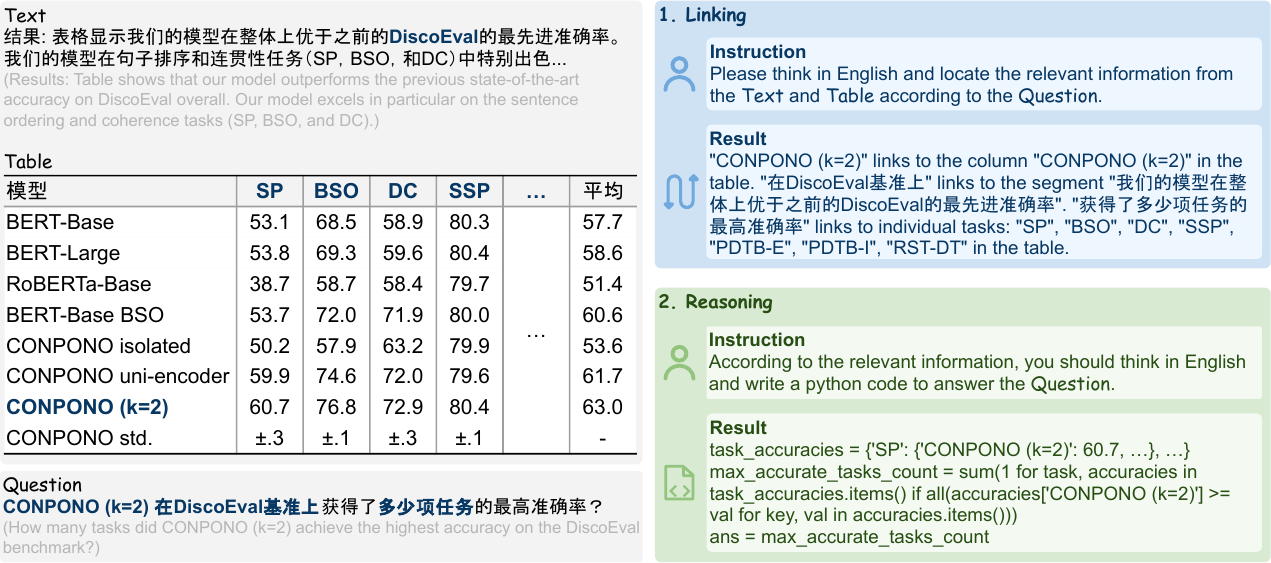}
    \caption{
    The overview of \ourmethod, which includes two modules:
    (\emph{i})~\textbf{Linking}: Mapping the entities in the question to the relevant information in tables or text, which are marked with \datatext{\textbf{blue}} in the left part.
    (\emph{ii})~\textbf{Reasoning}: Generating programs to solve the question using the information.
    We take the Chinese TATQA input as an example, with the corresponding English text provided in \colorbox{gray_light}{\textgray{(gray)}}.
    }
    \label{fig:method}
\end{figure*}

%% file: tab/answer_statistics.tex
\begin{tabular}{@{}l l l l c c c c@{}}
\toprule
\multirow{2}{*}{\textbf{Dataset}} & \multirow{2}{*}{\textbf{Domain}} & \multirow{2}{*}{\textbf{Scale}} & \multirow{2}{*}{\textbf{Answer Type}} & \multicolumn{3}{c}{\textbf{Answer Source}} & \multirow{2}{*}{\textbf{Total}}\\ 
\cmidrule(lr){5-7}
 & & & & \textbf{Text} & \textbf{Table} & \textbf{Hybrid} \\
\midrule
HybridQA~\cite{chen-etal-2020-hybridqa} & Wikipedia & $50$ & Span & $0$ & $0$ & $50$ & $50$ \\
\midrule
\multirow{3}{*}{TAT-QA~\cite{zhu-etal-2021-tat}} & \multirow{3}{*}{Finance} & \multirow{3}{*}{$100$} & Span & $10$ & $10$ & $20$ & $40$ \\ 
 & & & Arithmetic & $10$ & $10$ & $30$ & $50$ \\ 
 & & & Count & $2$ & $3$ & $5$ & $10$ \\
\midrule
\multirow{2}{*}{SciTAT~\cite{zhang2024scitat}} & \multirow{2}{*}{Science} & \multirow{2}{*}{$100$} & Span & $10$ & $20$ & $20$ & $50$ \\
 & & & Arithmetic & $10$ & $20$ & $20$ & $50$ \\
\midrule
Total & - & $250$ & - & $42$ & $63$ & $145$ & $250$ \\
\bottomrule
\end{tabular}

%% file: tex/3.methodology.tex
\subsection{Overview}
\ourmethod is designed to address the TATQA task under the multilingual setting. 
To align the strong TATQA capabilities of models in English with non-English languages, particularly low-resource languages, \ourmethod employs cross-lingual reasoning. 
To enable the model to perform English reasoning with non-English questions, tables, and text, \ourmethod is divided into two modules: Linking and Reasoning. 
As shown in Figure~\ref{fig:method}, Linking is responsible for locating relevant information from tables and text in the native language based on the question, and Reasoning performs reasoning in English based on the linked information.
The prompts used in \ourmethod are provided in Appendix~\ref{sec:prompt}.

\subsection{Linking}
Linking is used to map the entities in the question to the relevant information in the input text and tables so that Reasoning can directly utilize this information when generating the code. 
Specifically, we prompt the LLM to think in English and gradually map the relevant entities in the question to the information in the tables or text.

\subsection{Reasoning}
Reasoning is responsible for generating Python programs to solve the question and obtain the final answer based on the results of Linking. 
Considering that there are not only numbers in the answers, we also remind the LLM to note that the answers should be represented in the native language except for Arabic numerals.
Since the relevant information is extracted during Linking, Reasoning can directly use English variable names to define the numerical or tabular data when generating the program.

%% file: tex/4.experiment.tex
\subsection{Settings}

\begin{table*}[ht]
\centering
\tiny

\input{tab/main}
\caption{
EM (above) and F1 (below) of different models and baselines across languages on \ourdataset.
Avg. denotes the average performance of the baseline across all languages.
The best results of each model under each language are annotated in \textbf{bold}. 
}
\label{tab:main}
\end{table*}

\paragraph{Metrics}
We use Exact Match (EM) and F1 score to evaluate the answers, following prior works \cite{chen-etal-2020-hybridqa,zhu-etal-2021-tat}. 
EM refers to the proportion of predictions that exactly match the gold answer, and F1 measures the degree of overlap between the predicted and the gold answer in terms of their bag-of-words representation.

\paragraph{Models}
We evaluate \ourdataset using the open-source model Llama3.1-Instruct (Llama3.1)~\cite{dubey-etal-2024-llama3.1} and the closed-source model \texttt{gpt-4o}~\cite{openai2024gpt4technicalreport}. 
Llama3.1 is currently one of the best-performing open-source models, and \texttt{gpt-4o} is considered one of the leading closed-source models.

\paragraph{Baselines}
We compare \ourmethod with the following baselines with three-shot prompts, following previous works \cite{shi2023MGSM,li-etal-2024-eliciting-Multilingual-code}.
\begin{itemize}
    \item Native-CoT: solving the question using CoT~\cite{wei2022chain-of-thought} in the native language
    \item En-CoT: solving the question using CoT in English
    \item Native-PoT: prompting the LLM to generate code in the native language \cite{pal,pot}
    \item En-PoT: prompting the LLM to generate code in English 
    \item Three-Agent~\cite{fatemi2024three-agent} is the state-of-the-art method on the TAT-QA dataset. It consists of three agents: the analyst agent extracts relevant data and performs computations, and two critic agents evaluate the correctness of extraction and computation, respectively, and refine the results accordingly. 
    Due to computational resource limitations, we do not evaluate the performance of Three-Agent on \ourdataset using \texttt{gpt-4o}.
\end{itemize}
We present prompts for baselines and \ourmethod in Appendix~\ref{sec:prompt}.
Additionally, we provide results for both directly answering the question and reasoning after translating the input into English in Appendix~\ref{subsec:otherbaselines}.

\subsection{Main Experiments}
\label{subsec:main experiments}
A comparison of \ourmethod with other baselines across different languages is presented in Table~\ref{tab:main}. 
We observe that:
(\emph{i})~The performance on \ourdataset in non-English languages shows an average decrease of $19.4\%$ compared to English, underscoring the necessity of \ourdataset.
(\emph{ii})~\ourmethod demonstrates an average improvement of $3.3$ on EM and F1 over other baselines, reducing the performance gap between different languages by $23.2\%$, which validates the effectiveness.
(\emph{iii})~Despite these improvements, the EM and F1 of all baselines remain below $40$, highlighting the challenges of \ourdataset.

\paragraph{Baselines}
(\emph{i})~\ourmethod consistently outperforms Three-Agent because Three-Agent is not fully suited to HybridQA, which does not require computations \cite{chen-etal-2020-hybridqa}, or SciTAT, which involves complex calculations that are challenging to the inherent capabilities of models \cite{zhang2024scitat}. 
Additionally, the performance of multi-agent declines in non-English languages \cite{beyer2024clembench,chen-etal-2024-Cross-Agent}.
(\emph{ii})~The performance difference between reasoning in the native language and English is minimal. 
Although LLMs demonstrate stronger reasoning capabilities in English, the TATQA, compared to other tasks, relies more heavily on the capabilities of linking information, which presents greater challenges in cross-lingual reasoning \cite{min-etal-2019-cspider}. 
Therefore, \ourmethod mitigates this challenge, leading to improved performance. 
(\emph{iii})~PoT consistently outperforms CoT because numerical reasoning questions constitute a significant proportion of \ourdataset (see Table~\ref{tab:answer_statistics}), making PoT more suitable for solving these questions \cite{pot,zhao-etal-2024-docmath}.

\paragraph{Languages}
The models generally exhibit high performance on high-resource languages, such as English, German, Spanish, French, Russian, and Chinese, while their performance on low-resource languages tends to be poor. 
Moreover, models with stronger multilingual capabilities show smaller performance gaps across languages, with \texttt{gpt-4o} demonstrating the highest performance. 
This also underscores the necessity of evaluating multilingual performance on challenging tasks.


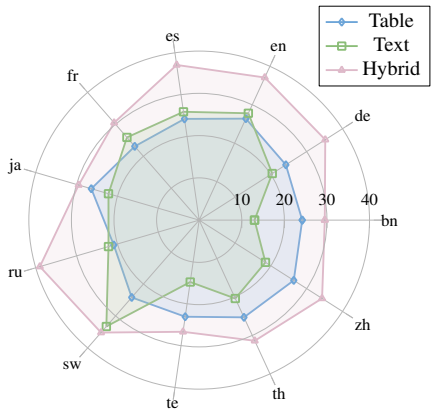
\begin{figure}[t]
    \centering
    \input{fig/answer_source}
    \caption{
        The EM of \ourmethod across different answer sources on \ourdataset using Llama3.1-70B.
    }
    \label{fig:answer_source}
\end{figure}

\paragraph{Answer Source}
We analyze the performance of \ourmethod using Llama3.1-70B across different answer sources, as shown in Figure~\ref{fig:answer_source}. 
The performance with other models and baselines across answer sources is provided in Appendix~\ref{subsec:appendix_answer_source}. 
The results show that:
(\emph{i})~The performance of the hybrid answer source generally outperforms those with a single answer source. 
Since \ourmethod, compared to other baselines (see Figure~\ref{fig:answer_sources_other_baselines}), enhances the links between the question and the context, integrating hybrid contextual information and alleviating the challenge.
(\emph{ii})~The performance across answer sources is influenced not only by the availability of language-specific resources but also by the characteristics of the language.
For instance, languages with complex morphological structures, such as German and Russian, perform worse when the answer source is text. 
In contrast, Swahili shows the highest performance on text-based sources, as its simpler morphology allows for easier linking of entities in the text to those in question \cite{tuan-nguyen-etal-2020-Vietnamese,zhang-etal-2023-xsemplr}.


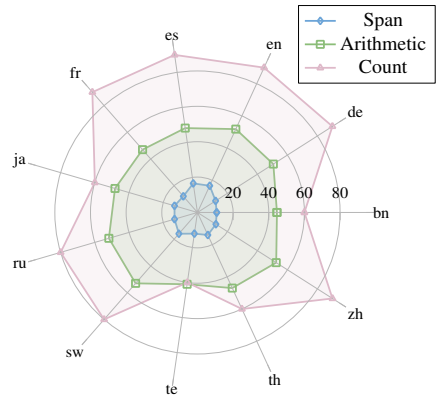
\begin{figure}[t]
    \centering
    \input{fig/answer_type}
    \vspace{-0.5em}
    \caption{
        The EM of \ourmethod across different answer types on \ourdataset using Llama3.1-70B.
    }
    \label{fig:answer_type}
    \vspace{-1em}
\end{figure}

\paragraph{Answer Type}
We compare the performance of \ourmethod using Llama3.1-70B on different answer types, as shown in Figure~\ref{fig:answer_type}. 
Results of other models and baselines across answer types are provided in Appendix~\ref{subsec:appendix_answer_types}. 
We observe that:
(\emph{i})~The model performs best on the Count type. 
This is because Span answers require extracting short phrases or summarizing conclusions from tables and text, making them more sensitive to word composition and order. 
Additionally, Arithmetic answers involve more complex computations than Count answers.
(\emph{ii})~The model performs better on high-resource languages than low-resource languages across answer types overall. 
Although \ourmethod narrows the performance gap, there remains a significant difference between high-resource and low-resource languages for all answer types.

\subsection{Analysis}

\begin{table*}[ht]
\centering
\tiny

\input{tab/prompt_language}

\caption{
EM (above) and F1 (below) of \ourmethod using the instructions and demonstrations of different languages on Llama3.1-70B.
The best results under each language are annotated in \textbf{bold}. 
Demo refers to demonstrations. 
Multi refers to demonstrations composed of multiple languages (English, Spanish, and Chinese).
Avg. denotes the average performance of the baseline across all languages.
}
\label{tab:prompt_language}
\end{table*}

\subsubsection{How does the Prompt Language Affect \ourmethod?}
We analyze the impact of using instructions and demonstrations in different languages on the performance of \ourmethod, as shown in Table~\ref{tab:prompt_language}. 
For the multilingual demonstrations, we select one demonstration each from English, Spanish, and Chinese, as the models perform well on these three high-resource languages, which also cover two language families. 
The English instruction and English demonstrations are the settings of \ourmethod used in the main experiments. 
The results indicate that:

(\emph{i})~Using English instructions generally outperforms using native instructions.
(\emph{ii})~Multilingual demonstrations outperform both native language and English demonstrations, suggesting that when sufficient native demonstrations are not available on the TATQA task, using demonstrations from the same language family or high-resource languages can also enhance performance.
Additionally, Swahili achieves the highest performance when using instructions and examples in the native language, highlighting its uniqueness. 

\begin{figure}[t]
    \centering
    \includegraphics[width=0.95\linewidth]{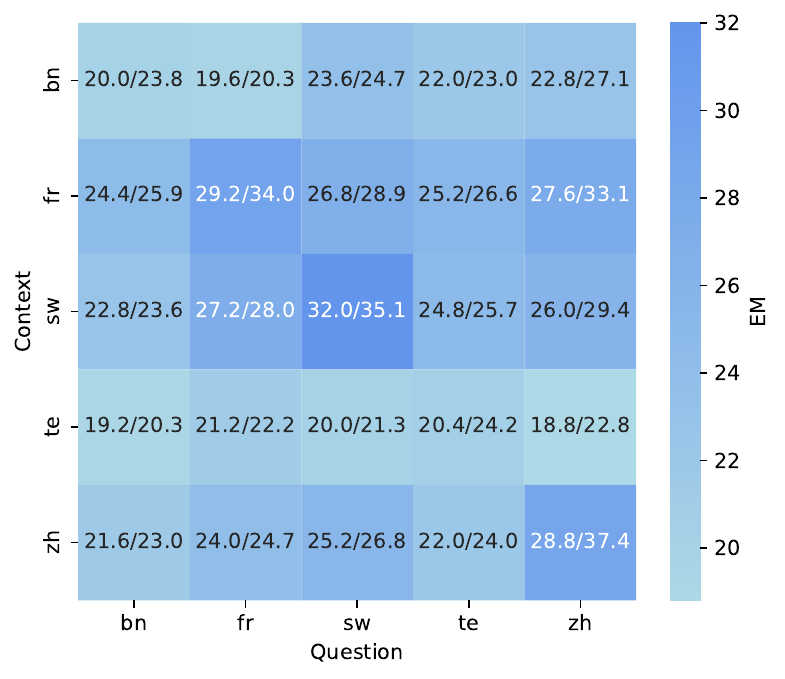}
    \caption{
    The EM/F1 of \ourmethod with questions and context (table and text) of different languages on \ourdataset using Llama3.1-70B. 
    }
    \label{fig:heatmap_cross}
\end{figure}

\subsubsection{How does the Language Affect \ourmethod in the Cross-lingual Setting?}
We evaluate the performance of \ourmethod in the cross-lingual setting, where the languages of the question and context are inconsistent, with results in Figure~\ref{fig:heatmap_cross}. 
We select high-resource languages (French and Chinese), and low-resource languages (Bengali, Swahili, and Telugu), covering $4$ language families. 
Our findings include:
(\emph{i})~Generally, \ourmethod shows improved performance when transitioning from low-resource to high-resource languages, while the opposite results in a decline. 
For instance, the performances on the French context with French and Chinese questions are relatively high, whereas the performances with three low-resource languages are lower.
(\emph{ii})~The model achieves the best performance when the question and context are both Swahili. 
This can be attributed to its relatively regular grammatical and lexical structures, which provide advantages when linking related information.


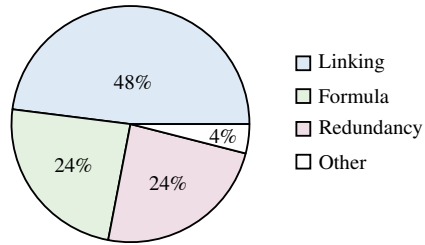
\begin{figure}[t]
    \centering
    \input{fig/error}
    \caption{
    The error types and their proportion of non-English performance in \ourmethod are inferior compared with English. 
    \textbf{Linking} refers to mapping entities in the question with incorrect information in the table or text. 
    \textbf{Formula} refers to using an incorrect formula. 
    \textbf{Redundancy} refers to outputting irrelevant information beyond the correct answer. 
    }
    \label{fig:error}
    \vspace{-1em}
\end{figure}

\subsection{Error Analysis}
\label{subsec:Error Analysis}
We analyze the reasons for the inferior performance of \ourmethod on non-English languages compared to English, as shown in Figure~\ref{fig:error}. 
Specifically, we select instances where \ourmethod achieved an EM of $1$ in English using Llama3.1-70B, but an EM of $0$ in non-English languages. 
For each language, we randomly sample five instances, with a total of $50$ errors for comparative analysis. 
Examples of errors corresponding to each type are provided in Appendix~\ref{subsec:case study}. 
Below, we present a detailed discussion of each error type:

(\emph{i})~\textbf{Linking}: 
Due to the relatively weaker abilities in non-English languages compared to English, even though \ourmethod initially prompts the model to focus on linking, the model still faces significant challenges in linking. 
These challenges are particularly pronounced in languages with complex orthographies, such as Japanese (with its hiragana and katakana scripts), or morphologically rich languages like French and German.
(\emph{ii})~\textbf{Formula} 
highlights the gap in the numerical reasoning abilities between non-English languages and English.
(\emph{iii})~\textbf{Redundancy}  
reflects the relatively weaker ability of instruction-following.

In summary, the inferior performance on non-English languages and the specific properties of languages leads to the lower performance of \ourmethod on non-English languages, which also demonstrates the necessity of \ourdataset.

%% file: tab/main.tex
\begin{tabular}{l|l|ccccccccccc|c}
    \toprule
    \textbf{Model} & \textbf{Method} & \textbf{bn} & \textbf{de} & \textbf{en} & \textbf{es} & \textbf{fr} & \textbf{ja} & \textbf{ru} & \textbf{sw} & \textbf{te} & \textbf{th} & \textbf{zh} & \textbf{Avg.} \\
    \midrule
    \multirow{5}{*}{Llama3.1-8B} 
    & Native-CoT & $11.2$ & $14.0$ & $20.8$ & $12.8$ & $8.0$ & $13.2$ & $15.2$ & $9.2$ & $12.4$ & $12.0$ & $13.6$ & $12.9$ \\
    & En-CoT    & $10.8$ & $14.6$ & $20.8$ & $12.4$ & $8.4$ & $13.2$ & $15.2$ & $9.2$ & $12.0$ & $12.0$ & $13.6$ & $12.9$ \\
    & Native-PoT & $18.0$ & $18.4$ & $21.2$ & $22.8$ & $18.4$ & $19.6$ & $22.8$ & $17.2$ & $6.8$ & $21.2$ & $19.6$ & $18.7$ \\
    & En-PoT    & $13.6$ & $12.8$ & $21.2$ & $20.8$ & $14.4$ & $20.8$ & $20.0$ & $10.4$ & $7.6$ & $19.2$ & $19.6$ & $16.6$ \\
    & Three-Agent & $10.0$ & $16.0$ & $21.6$ & $20.8$ & $15.6$ & $13.6$ & $12.0$ & $13.2$ & $9.2$ & $15.2$ & $18.4$ & $15.1$ \\
    & \ourmethod      & \bm{$20.0$} & \bm{$22.4$} & \bm{$27.6$} & \bm{$25.6$} & \bm{$20.0$} & \bm{$25.6$} & \bm{$25.2$} & \bm{$17.2$} & \bm{$14.4$} & \bm{$22.8$} & \bm{$23.6$} & \bm{$22.2$} \\
    \midrule
    \multirow{5}{*}{Llama3.1-70B} 
    & Native-CoT & $18.8$ & $20.8$ & $25.6$ & $23.6$ & $24.8$ & $22.4$ & $25.2$ & $23.6$ & $18.8$ & $21.6$ & $21.6$ & $22.4$ \\
    & En-CoT    & $18.4$ & $19.6$ & $25.6$ & $23.6$ & $20.0$ & $22.0$ & $25.2$ & $24.0$ & $19.6$ & $22.4$ & $22.0$ & $22.0$ \\
    & Native-PoT & $22.8$ & $24.4$ & $30.4$ & $28.4$ & $26.4$ & $18.4$ & $28.0$ & $28.4$ & $22.0$ & $26.0$ & $22.0$ & $25.2$ \\
    & En-PoT    & $23.6$ & $26.0$ & $30.4$ & $27.6$ & $26.4$ & $25.6$ & $28.4$ & $26.4$ & $22.0$ & $25.2$ & $26.8$ & $26.2$ \\
    & Three-Agent & $16.0$ & $25.6$ & $29.2$ & $23.6$ & $22.0$ & $25.6$ & $20.8$ & $22.4$ & $20.0$ & $19.6$ & $23.6$ & $22.6$ \\
    & \ourmethod      & \bm{$24.0$} & \bm{$28.0$} & \bm{$31.2$} & \bm{$29.2$} & \bm{$26.8$} & \bm{$26.8$} & \bm{$28.8$} & \bm{$30.8$} & \bm{$22.8$} & \bm{$26.8$} & \bm{$28.0$} & \bm{$27.6$} \\
    \midrule
    \multirow{5}{*}{\texttt{gpt-4o}} 
    & Native-CoT & $21.2$ & $27.2$ & $31.2$ & $26.8$ & $23.6$ & $19.2$ & $24.8$ & $24.8$ & $26.8$ & $26.8$ & $24.4$ & $24.7$ \\
    & En-CoT    & $23.6$ & $24.8$ & $31.2$ & $26.0$ & $22.0$ & $26.4$ & $26.4$ & $28.0$ & $22.0$ & $23.2$ & $24.8$ & $25.3$ \\
    & Native-PoT & $24.4$ & $30.4$ & $30.0$ & $30.4$ & $26.4$ & $21.2$ & $27.2$ & $26.4$ & $26.8$ & $24.8$ & $28.0$ & $27.6$ \\
    & En-PoT    & $24.0$ & $24.4$ & $30.0$ & $30.0$ & $26.4$ & $21.2$ & $27.2$ & $26.4$ & $21.2$ & $27.2$ & $24.4$ & $26.2$ \\
    & \ourmethod      & \bm{$30.0$} & \bm{$32.4$} & \bm{$35.2$} & \bm{$32.4$} & \bm{$29.6$} & \bm{$28.8$} & \bm{$31.2$} & \bm{$31.2$} & \bm{$30.8$} & \bm{$30.4$} & \bm{$30.9$} & \bm{$31.1$} \\
    \bottomrule
\end{tabular}

\begin{tabular}{l|l|ccccccccccc|c}
    \toprule
    \textbf{Model} & \textbf{Method} & \textbf{bn} & \textbf{de} & \textbf{en} & \textbf{es} & \textbf{fr} & \textbf{ja} & \textbf{ru} & \textbf{sw} & \textbf{te} & \textbf{th} & \textbf{zh} & \textbf{Avg.} \\
    \midrule
    \multirow{5}{*}{Llama3.1-8B} 
    & Native-CoT & $13.2$ & $16.1$ & $23.7$ & $17.2$ & $11.2$ & $14.5$ & $17.3$ & $14.0$ & $14.9$ & $14.6$ & $21.5$ & $16.2$ \\
    & En-CoT    & $13.4$ & $16.6$ & $23.7$ & $17.9$ & $12.4$ & $15.2$ & $17.8$ & $14.0$ & $14.9$ & $14.9$ & $22.7$ & $16.7$ \\
    & Native-PoT & $19.1$ & $18.9$ & $22.8$ & $24.2$ & $19.3$ & $19.9$ & $23.1$ & $17.8$ & $6.9$ & $22.4$ & $21.7$ & $19.6$ \\
    & En-PoT    & $14.1$ & $13.7$ & $22.8$ & $21.3$ & $15.1$ & $21.5$ & $20.6$ & $11.0$ & $7.8$ & $20.1$ & $21.7$ & $17.4$ \\
    & Three-Agent & $15.7$ & $20.5$ & $26.4$ & $25.8$ & $20.6$ & $15.1$ & $16.0$ & $17.4$ & $13.9$ & $18.8$ & $26.1$ & $19.7$ \\
    & \ourmethod      & \bm{$21.3$} & \bm{$24.2$} & \bm{$31.9$} & \bm{$27.8$} & \bm{$22.4$} & \bm{$26.1$} & \bm{$27.0$} & \bm{$20.0$} & \bm{$15.2$} & \bm{$24.6$} & \bm{$28.0$} & \bm{$24.4$} \\
    \midrule
    \multirow{5}{*}{Llama3.1-70B} 
    & Native-CoT & $21.6$ & $22.8$ & $29.3$ & $27.0$ & $28.1$ & $24.4$ & $27.3$ & $26.6$ & $21.3$ & $24.0$ & $28.3$ & $25.5$ \\
    & En-CoT    & $21.6$ & $22.4$ & $29.3$ & $27.9$ & $23.6$ & $24.7$ & $27.7$ & $27.3$ & $22.3$ & $26.3$ & $29.4$ & $25.7$ \\
    & Native-PoT & $24.8$ & $26.2$ & $32.9$ & $30.6$ & $29.0$ & $18.7$ & $29.4$ & $29.9$ & $24.0$ & $28.4$ & $30.0$ & $27.0$ \\
    & En-PoT    & $25.8$ & $27.9$ & $32.9$ & $30.2$ & $28.7$ & $27.2$ & $30.3$ & $28.7$ & $25.0$ & $27.3$ & $30.9$ & $28.5$ \\
    & Three-Agent & $22.2$ & $30.8$ & $34.5$ & $31.3$ & $28.4$ & $28.2$ & $25.5$ & $27.1$ & $24.3$ & $24.8$ & $33.3$ & $28.2$ \\
    & \ourmethod      & \bm{$26.3$} & \bm{$31.3$} & \bm{$35.3$} & \bm{$34.6$} & \bm{$31.1$} & \bm{$29.4$} & \bm{$33.5$} & \bm{$34.7$} & \bm{$25.9$} & \bm{$30.5$} & \bm{$34.9$} & \bm{$31.6$} \\
    \midrule
    \multirow{5}{*}{\texttt{gpt-4o}} 
    & Native-CoT & $27.0$ & $33.8$ & $38.8$ & $36.3$ & $30.2$ & $21.8$ & $31.9$ & $31.3$ & $31.3$ & $30.9$ & $38.2$ & $31.6$ \\
    & En-CoT    & $28.0$ & $32.1$ & $38.8$ & $33.1$ & $27.2$ & $28.8$ & $32.4$ & $33.6$ & $25.0$ & $28.8$ & $34.6$ & $31.1$ \\
    & Native-PoT & $26.7$ & $33.3$ & $32.5$ & $32.5$ & $28.7$ & $22.5$ & $29.9$ & $27.7$ & $29.4$ & $27.2$ & $29.5$ & $30.1$ \\
    & En-PoT    & $26.2$ & $26.8$ & $31.3$ & $32.5$ & $28.7$ & $22.5$ & $29.9$ & $27.7$ & $25.0$ & $29.0$ & $27.2$ & $28.0$ \\
    & \ourmethod      & \bm{$32.9$} & \bm{$35.5$} & \bm{$38.9$} & \bm{$35.7$} & \bm{$32.5$} & \bm{$32.1$} & \bm{$33.1$} & \bm{$34.0$} & \bm{$34.7$} & \bm{$35.1$} & \bm{$34.5$} & \bm{$34.7$} \\
    \bottomrule
\end{tabular}

%% file: fig/answer_source.tex
\resizebox{0.75\linewidth}{!}{
\begin{tikzpicture}
\begin{polaraxis}[
    title={\empty},
    xlabel={\empty},
    ylabel={\empty},
    xtick={0,32.73,65.46,98.19,130.92,163.65,196.38,229.11,261.84,294.57,327.30},
    xticklabels={bn, de, en, es, fr, ja, ru, sw, te, th, zh},
    ytick={10,20,30,40},
    yticklabels={10,20,30,40},
    grid=both,
    tick label style={font=\small},
    major grid style={solid, gray}, 
    minor grid style={solid, gray}, 
    axis line style={draw=none},
    axis on top=true,                
    grid style={gray},              
    line join=bevel,
    tick align=outside,
]

\addplot[
    color=data_blue!300,
    mark=diamond,
    style={line width=1pt},
    fill=data_blue!250,fill opacity=0.2
] coordinates {
    (0,24.2)(32.73,24.2)(65.46,26.4)(98.19,24.2)(130.92,23.1)(163.65,26.4)(196.38,20.9)(229.11,24.2)(261.84,23.1)(294.57,25.3)(327.30,26.4)(0,24.2)
} -- cycle;

\addplot[
    color=reasoner_green!300,
    mark=square,
    style={line width=1pt},
    fill=reasoner_green!250,fill opacity=0.2
] coordinates {
    (0,13.0)(32.73,20.4)(65.46,27.8)(98.19,25.9)(130.92,25.9)(163.65,22.2)(196.38,22.2)(229.11,33.3)(261.84,14.8)(294.57,20.4)(327.30,18.5)(0,13.0)
} -- cycle;

\addplot[
    color=annotator_pink!150,
    mark=triangle,
    style={line width=1pt},
    fill=annotator_pink,fill opacity=0.2
] coordinates {
    (0,29.5)(32.73,35.2)(65.46,37.1)(98.19,37.1)(130.92,30.5)(163.65,29.5)(196.38,39.0)(229.11,35.2)(261.84,26.7)(294.57,31.4)(327.30,34.3)(0,29.5)
} -- cycle;

\legend{Table, Text, Hybrid}

\end{polaraxis}
\end{tikzpicture}
}

%% file: fig/answer_type.tex
\resizebox{0.75\linewidth}{!}{
\begin{tikzpicture}
\begin{polaraxis}[
    title={\empty},
    xlabel={\empty},
    ylabel={\empty},
    xtick={0,32.73,65.46,98.19,130.92,163.65,196.38,229.11,261.84,294.57,327.30},
    xticklabels={bn, de, en, es, fr, ja, ru, sw, te, th, zh},
    ytick={20,40,60,80,100},
    yticklabels={20,40,60,80,100},
    grid=both,
    tick label style={font=\small},
    major grid style={solid, gray}, 
    minor grid style={solid, gray}, 
    axis line style={draw=none},
    axis on top=true,                
    grid style={gray},              
    line join=bevel,
    tick align=outside,
]

\addplot[
    color=data_blue!300,
    mark=diamond,
    style={line width=1pt},
    fill=data_blue!250,fill opacity=0.2
] coordinates {
    (0,10.8)(32.73,12.1)(65.46,16.6)(98.19,16.6)(130.92,12.1)(163.65,13.4)(196.38,13.4)(229.11,15.9)(261.84,12.1)(294.57,14.0)(327.30,12.3)(0,10.8)
} -- cycle;

\addplot[
    color=reasoner_green!300,
    mark=square,
    style={line width=1pt},
    fill=reasoner_green!250,fill opacity=0.2
] coordinates {
    (0,44.6)(32.73,50.6)(65.46,51.8)(98.19,48.2)(130.92,47.0)(163.65,48.2)(196.38,51.8)(229.11,53.0)(261.84,41.0)(294.57,47.0)(327.30,52.5)(0,44.6)
} -- cycle;

\addplot[
    color=annotator_pink!150,
    mark=triangle,
    style={line width=1pt},
    fill=annotator_pink,fill opacity=0.2
] coordinates {
    (0,60.0)(32.73,90.0)(65.46,90.0)(98.19,90.0)(130.92,90.0)(163.65,60.0)(196.38,80.0)(229.11,80.0)(261.84,40.0)(294.57,60.0)(327.30,90.0)(0,60.0)
} -- cycle;

\legend{Span, Arithmetic, Count}

\end{polaraxis}
\end{tikzpicture}
}

%% file: tab/prompt_language.tex
\begin{tabular}{l|l|ccccccccccc|c}
\toprule
\textbf{Instruction} & \textbf{Demo} & \textbf{bn} & \textbf{de} & \textbf{en} & \textbf{es} & \textbf{fr} & \textbf{ja} & \textbf{ru} & \textbf{sw} & \textbf{te} & \textbf{th} & \textbf{zh} & \textbf{Avg.} \\
\midrule
\multirow{3}{*}{Native} & Native & $20.0$ & $28.4$ & $28.4$ & $29.2$ & $29.2$ & $27.6$ & $27.6$ & \bm{$32.0$} & $20.4$ & $25.2$ & \bm{$28.8$} & $27.0$ \\
& Multi & $22.0$ & \bm{$30.0$} & $30.4$ & $30.4$ & $28.4$ & $26.0$ & $26.4$ & $28.8$ & $24.4$ & $24.4$ & $24.8$ & $26.9$ \\
& En & $20.8$ & $29.2$ & $28.4$ & $24.8$ & $27.2$ & $24.0$ & $28.4$ & $29.2$ & $19.6$ & $21.2$ & $24.4$ & $24.9$ \\
\midrule
\multirow{3}{*}{En} & Native & \bm{$27.6$} & $26.8$ & $28.4$ & $29.6$ & $25.2$ & $25.6$ & $29.2$ & $30.0$ & $26.0$ & \bm{$28.0$} & $26.8$ & $27.6$ \\
& Multi & $26.4$ & $27.2$ & $30.4$ & \bm{$30.8$} & \bm{$29.6$} & \bm{$29.2$} & \bm{$30.0$} & $30.0$ & \bm{$27.2$} & $27.2$ & \bm{$28.8$} & \bm{$28.8$} \\
& En & $24.0$ & $28.0$ & \bm{$31.2$} & $29.2$ & $26.8$ & $26.8$ & $28.8$ & $30.8$ & $22.8$ & $26.8$ & $28.0$ & $27.6$ \\
\bottomrule
\end{tabular}

\begin{tabular}{l|l|ccccccccccc|c}
\toprule
\textbf{Instruction} & \textbf{Demo} & \textbf{bn} & \textbf{de} & \textbf{en} & \textbf{es} & \textbf{fr} & \textbf{ja} & \textbf{ru} & \textbf{sw} & \textbf{te} & \textbf{th} & \textbf{zh} & \textbf{Avg.} \\
\midrule
\multirow{3}{*}{Native} & Native & $23.8$ & \bm{$33.9$} & $33.8$ & \bm{$35.8$} & $\bm{34.0}$ & $30.1$ & $31.7$ & \bm{$35.1$} & $24.2$ & $28.3$ & \bm{$37.4$} & $31.7$ \\
& Multi & $24.6$ & $32.3$ & $\bm{35.4}$ & $35.0$ & $31.6$ & $27.6$ & $28.8$ & $30.7$ & $26.3$ & $26.7$ & $30.7$ & $30.0$ \\
& En & $24.4$ & $33.6$ & $33.8$ & $30.2$ & $32.0$ & $22.8$ & $31.8$ & $31.6$ & $22.3$ & $23.5$ & $30.7$ & $28.8$ \\
\midrule
\multirow{3}{*}{En} & Native & \bm{$30.5$} & $30.3$ & $33.8$ & $32.7$ & $29.6$ & $28.5$ & $33.0$ & $33.6$ & $28.8$ & \bm{$31.4$} & $34.1$ & $31.5$ \\
& Multi & $28.9$ & $29.9$ & $\bm{35.4}$ & $34.1$ & $32.2$ & $\bm{31.7}$ & $32.6$ & $33.0$ & \bm{$29.8$} & $31.2$ & $34.7$ & \bm{$32.1$} \\
& En & $26.3$ & $31.3$ & $35.3$ & $34.6$ & $31.1$ & $29.4$ & $\bm{33.5}$ & $34.7$ & $25.9$ & $30.5$ & $34.9$ & $31.6$ \\
\bottomrule
\end{tabular}

%% file: fig/error.tex
\resizebox{0.75\linewidth}{!}{
    \begin{tikzpicture}[scale=0.6]
        \small
        \pie[
            pos={8,0},
            color={data_blue!70, reasoner_green!70, annotator_pink!70, white},
            text=legend
        ]{
            48/Linking,
            24/Formula,
            24/Redundancy,
            4/Other
        }
    \end{tikzpicture}
}

%% file: tex/5.related.tex
\subsection{Multilingual Datasets}
To evaluate the performance of models across different languages, several multilingual datasets have been proposed for different tasks, such as question answering \cite{liu-etal-2019-xqa,clark-etal-2020-tydiQA,longpre-etal-2021-mkqa}, natural language inference \cite{conneau-etal-2018-xnli}, text summarization \cite{giannakopoulos-etal-2015-multiling-Summarization,ladhak-etal-2020-wikilingua,scialom-etal-2020-mlsum}, numerical reasoning \cite{shi2023MGSM}, code generation \cite{peng-etal-2024-humanevalxl}, text-to-SQL \cite{MultiSpider}, and readability \cite{trokhymovych-etal-2024-open-Readability,naous-etal-2024-readme}, among others. 
Additionally, numerous multilingual datasets have been collected for different tasks \cite{hu-2020-XTREME,ruder-etal-2021-xtremer,zhang2024pmmeval-multitask,singh-etal-2024-indicgenbench}. 
However, to date, there is no multilingual TATQA dataset, resulting in a lack of evaluation and analysis of multilingual TATQA capabilities and a gap with real scenarios. 
Therefore, we introduce \ourdataset, a multilingual TATQA dataset, and provide a detailed analysis of the challenges in multilingual TATQA.

\subsection{QA Datasets for the Table and Text}
Currently, QA datasets for the table and text primarily focus on a single language. 
For instance, HybridQA~\cite{chen-etal-2020-hybridqa} collects English tables and associated text from Wikipedia.
TAT-QA~\cite{zhu-etal-2021-tat}, FinQA~\cite{chen-etal-2021-finqa}, DOCMATH-EVAL~\cite{zhao-etal-2024-docmath}, and FinanceMATH~\cite{zhao-etal-2024-FinanceMATH} focus on numerical computation in the financial domain, and SciTAT~\cite{zhang2024scitat} addresses questions based on tables and text from English scientific papers. 
However, single-language datasets cannot evaluate the multilingual TATQA capabilities, and overlook the diverse languages in real scenarios. 
So we propose \ourdataset: the first multilingual TATQA dataset, involving $11$ languages and $8$ language families.
A comparison of \ourdataset and prior works is presented in Appendix~\ref{sec:comparison}.

The current works on enhancing TATQA performance primarily focus on retrieving relevant information from the context \cite{luo2023hrot,bardhan2024ttqars,glenn-etal-2024-blendsql} and generating programs, equations, or step-by-step reasoning process to derive the final answer \cite{tonglet-etal-2023-seer,TAT-LLM,fatemi2024three-agent}.
For example, S3HQA~\cite{lei-etal-2023-s3hqa} emphasizes retrieving, where a retriever is initially trained, followed by further filtering based on the question type. 
Hpropro~\cite{shi-etal-2024-hpropro} focuses on generating, providing LLMs with commonly used functions to facilitate direct invocation during code generation.
However, previous methods are designed for single-language scenarios, directly used to other languages could lead to performance degradation.
To address this, we propose \ourmethod, a multilingual baseline that aligns the English TATQA capabilities to other languages. 

%% file: tex/6.appendix.tex
\section{Comparison with Previous Datasets}
\label{sec:comparison}
In this section, we make a detailed comparison between \ourdataset and previous TATQA datasets, as shown in Table.
It can be seen that \ourdataset is the first multilingual TATQA dataset, and it gathers previous datasets from three mainstream fields.

\begin{table*}[ht]
    \centering
    \small
    \input{tab/comparison_tat}
    \caption{
        Comparison of \ourdataset to previous TATQA datasets.
    }
    \label{tab:comparison_tat}
\end{table*}

\section{Manual Annotation Process}

\subsection{Annotator Training Process}
\label{subsec:Annotator Training Process}
We hire graduate students majoring in Computer Science who are willing to participate in the annotation process. 
First, we provide annotators with a clear definition of the task, the specific checks and revisions required (as described in Section \S\ref{subsec:Rationale Annotation} and \S\ref{subsec:Instance Translation}), and instructions on how to use the annotation interface. 
The annotation interface is shown in \S\ref{subsec:Annotation Interface}. 
We also inform them of the annotation deadline and encourage them to discuss any uncertainties with us promptly.
Finally, a total of five annotators complete the annotation for \S\ref{subsec:Rationale Annotation} and \S\ref{subsec:Instance Translation}, with a combined time of one month.

\subsection{Annotation Interface}
\label{subsec:Annotation Interface}
\begin{figure*}
    \centering
    \includegraphics[width=.95\linewidth]{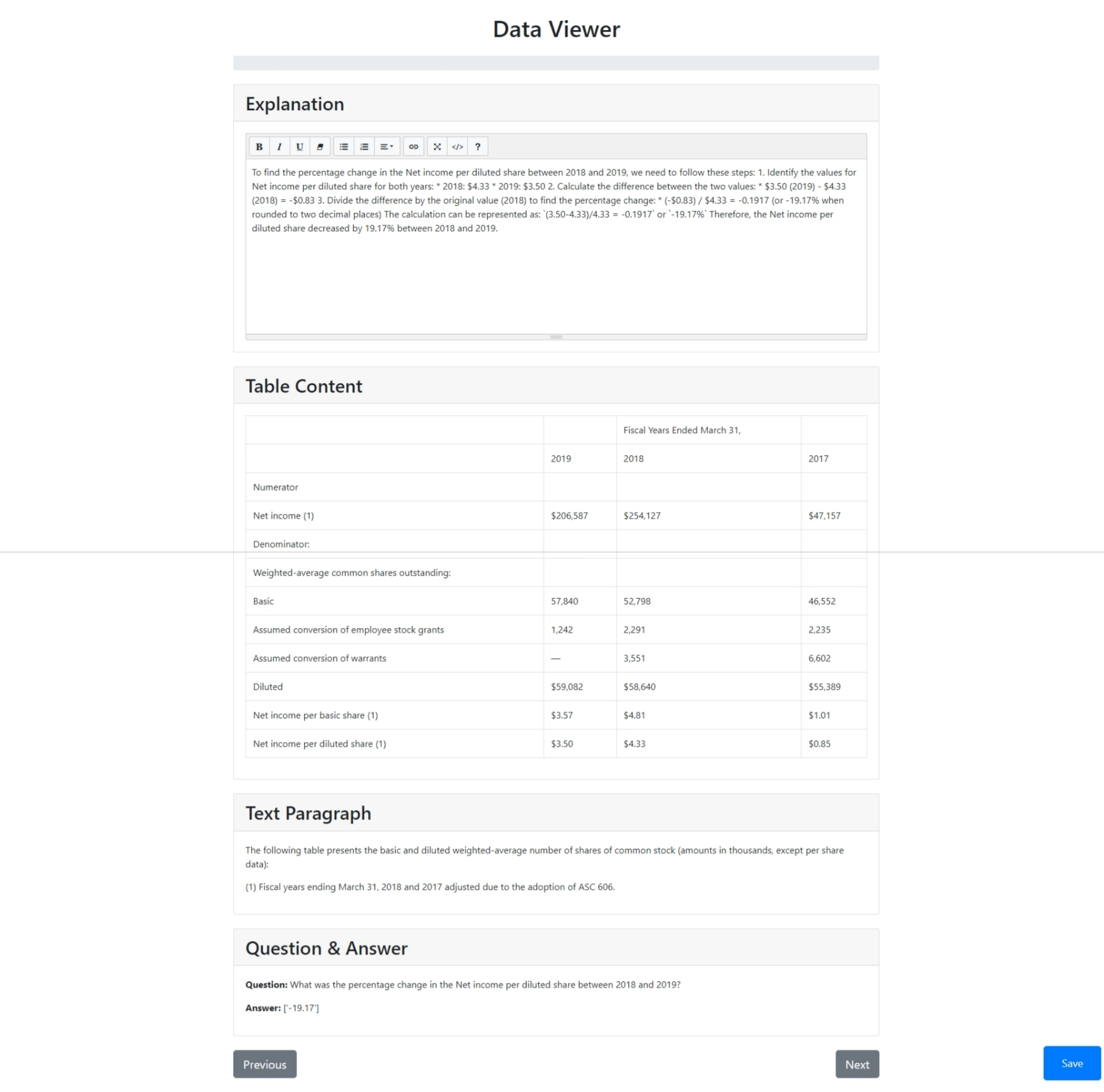}
    \vspace{-0.5em}
    \caption{
    The annotation interface is provided to annotators to check the accuracy of the generated rationales.
    }
    \label{fig:rationale_check_tool}
\end{figure*}

\begin{figure*}
    \centering
    \includegraphics[width=.95\linewidth]{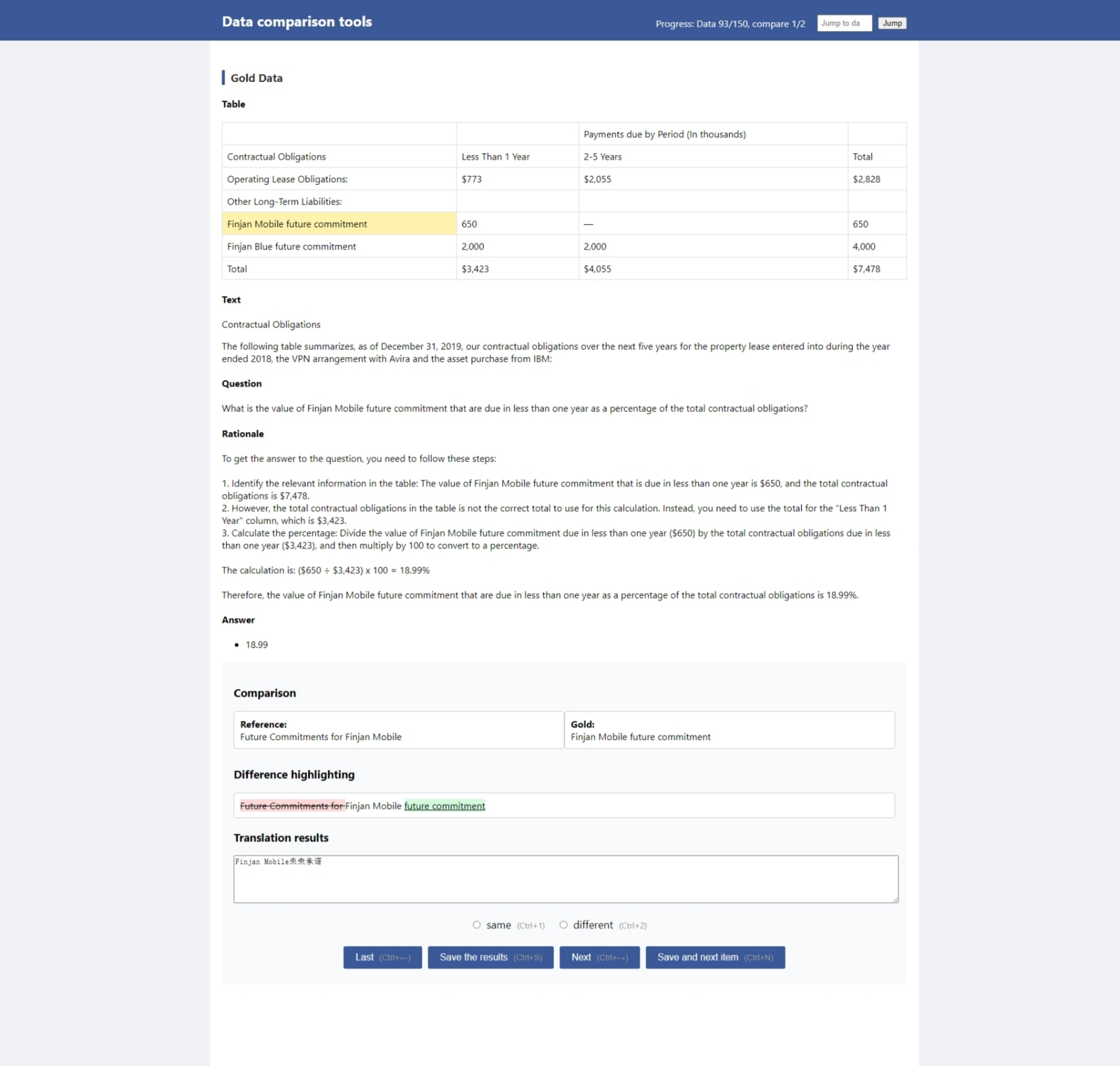}
    \vspace{-0.5em}
    \caption{
    The annotation interface is provided to annotators to check the consistency of the back translation and the original English instance and refine the translated instances.
    }
    \label{fig:translation_check_tool}
\end{figure*}
In this subsection, we show the interfaces annotated by the annotator, which are developed by ourselves, as shown in Figure~\ref{fig:rationale_check_tool} and Figure~\ref{fig:translation_check_tool}.



\section{Prompt}
\label{sec:prompt}
In this section, we show the prompts we use to conduct experiments.
Table~\ref{tab:prompt_baseline} and Table~\ref{tab:prompt_our_method} show the prompts of the baselines and \ourmethod in experiments respectively, with French as the example language.
The prompt of Three-Agent~\cite{fatemi2024three-agent} follows the prompt provided in the original paper.
We maintain the unity of demonstrations between different languages and baselines, as shown in Table~\ref{tab:prompt_our_method}.

\begin{table*}[ht]
    \centering
    \small
    \input{tab/prompt_baseline}
    \caption{
    The prompts of baselines for French.
    }
    \label{tab:prompt_baseline}
\end{table*}

\begin{table*}[ht]
    \centering
    \small
    \input{tab/prompt_our_method}
    \caption{
    The prompts of \ourmethod for French.
    }
    \label{tab:prompt_our_method}
\end{table*}

\section{Additional Experiments}
\label{sec:Additional Experiments}

\subsection{Other Baselines}
\label{subsec:otherbaselines}

\begin{table*}[ht]
\centering
\small

\input{tab/other_baselines}
\caption{
EM/F1 of different models and baselines across languages on \ourdataset.
The best results of each model under each language are annotated in \textbf{bold}. 
}
\label{tab:other_baselines}
\end{table*}

In this subsection, we show the results of directly answering the questions (Direct), solving the question with English CoT (Trans-CoT) and PoT (Trans-PoT) after translating the question and context (including the table and text) to English, as shown in Table~\ref{tab:other_baselines}.
\ourmethod consistently and significantly outperforms all baseline methods, demonstrating its effectiveness. 

Additionally, we observe the following:
(\emph{i})~Compared to direct question answering, the overall performance of Native-CoT, Native-PoT, En-CoT, and En-PoT shows substantial improvement (see Table~\ref{tab:main}).
(\emph{i})~The performance of Trans-CoT and Trans-PoT is unstable, primarily due to limitations in the quality of Google Translation. 
On the one hand, Google Translation struggles to maintain table formatting during translation, especially for low-resource languages such as Bengali and Swahili, leading to information loss \cite{MultiSpider}. 
On the other hand, when utilizing back-translation via Google Translation, token consistency with the original table or text cannot be guaranteed.

\subsection{Answer Sources}
\label{subsec:appendix_answer_source}

\begin{figure*}[t]
    \centering
    \begin{subfigure}[b]{0.48\linewidth}
        \centering
        \input{fig/answer_source_8b}
    \end{subfigure}
    \hfill 
    \begin{subfigure}[b]{0.48\linewidth}
        \centering
        \input{fig/answer_source_4o}
    \end{subfigure}
    \vspace{-1em}
    \caption{
        The left part is the EM of \ourmethod across different answer sources on \ourdataset using Llama3.1-8B.
        The right part is the EM of \ourmethod across different answer sources on \ourdataset using \texttt{gpt-4o}.
    } 
    \label{fig:answer_sources_other_models}
    \vspace{-1em}
\end{figure*}

\begin{figure*}[t]
    \centering
    \begin{subfigure}[b]{0.48\linewidth}
        \centering
        \input{fig/answer_source_en_CoT}
    \end{subfigure}
    \hfill 
    \begin{subfigure}[b]{0.48\linewidth}
        \centering
        \input{fig/answer_source_en_PoT}
    \end{subfigure}
    \vspace{-1em}
    \caption{
        The left part is the EM of En-CoT across different answer sources on \ourdataset using Llama3.1-70B.
        The right part is the EM of En-PoT across different answer sources on \ourdataset using Llama3.1-70B.
    } 
    \label{fig:answer_sources_other_baselines}
    \vspace{-1em}
\end{figure*}

In this subsection, we present the performance of different models and baselines on various answer sources in our dataset, as illustrated in Figure~\ref{fig:answer_sources_other_models} and Figure~\ref{fig:answer_sources_other_baselines}.
From Figure~\ref{fig:answer_sources_other_models}, it can be observed that multilingual models with better overall performance tend to exhibit smaller performance gaps across different languages. 
However, even \texttt{gpt-4o} still cannot entirely eliminate the discrepancies.
From Figure~\ref{fig:answer_sources_other_baselines}, in comparison with Figure~\ref{fig:answer_source}, \ourmethod demonstrates performance improvements across all answer sources, with a particularly significant enhancement for hybrid answer sources. 
This is attributed to the ability to better establish connections to relevant information of \ourmethod, thereby mitigating the challenges posed by the heterogeneity of answer sources.

\subsection{Answer Types}
\label{subsec:appendix_answer_types}

\begin{figure*}[t]
    \centering
    \begin{subfigure}[b]{0.48\linewidth}
        \centering
        \input{fig/answer_type_8b}
    \end{subfigure}
    \hfill 
    \begin{subfigure}[b]{0.48\linewidth}
        \centering
        \input{fig/answer_type_4o}
    \end{subfigure}
    \vspace{-1em}
    \caption{
        The left part is the EM of \ourmethod across different answer types on \ourdataset using Llama3.1-8B.
        The right part is the EM of \ourmethod across different answer types on \ourdataset using \texttt{gpt-4o}.
    } 
    \label{fig:answer_types_other_models}
    \vspace{-1em}
\end{figure*}

\begin{figure*}[t]
    \centering
    \begin{subfigure}[b]{0.48\linewidth}
        \centering
        \input{fig/answer_type_en_CoT}
    \end{subfigure}
    \hfill 
    \begin{subfigure}[b]{0.48\linewidth}
        \centering
        \input{fig/answer_type_en_PoT}
    \end{subfigure}
    \vspace{-1em}
    \caption{
        The left part is the EM of En-CoT across different answer types on \ourdataset using Llama3.1-70B.
        The right part is the EM of En-PoT across different answer types on \ourdataset using Llama3.1-70B.
    } 
    \label{fig:answer_types_other_baselines}
    \vspace{-1em}
\end{figure*}

In this subsection, we present the performance of different models and baselines across various answer types in \ourdataset, as illustrated in Figure~\ref{fig:answer_types_other_models} and Figure~\ref{fig:answer_types_other_baselines}.
As shown in Figure~\ref{fig:answer_types_other_models}, even for \texttt{gpt-4o}, the performance for high-resource languages is consistently superior to that for low-resource languages across different answer types.
Figure~\ref{fig:answer_types_other_baselines} demonstrates that, compared to Figure~\ref{fig:answer_type}, \ourmethod reduces the performance gap between languages of varying resource levels to some extent and uniformly improves performance across different answer types.

\subsection{Case Study}
\label{subsec:case study}

\begin{figure*}[t]
    \centering
    \includegraphics[width=.85\linewidth]{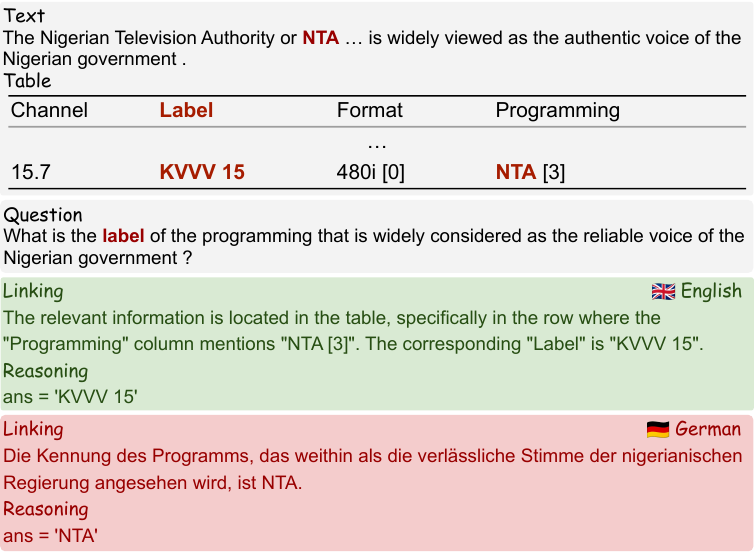}
    \caption{
    The case for the error type of "Linking".
    }
    \label{fig:case_linking}
\end{figure*}

\begin{figure*}[t]
    \centering
    \includegraphics[width=.85\linewidth]{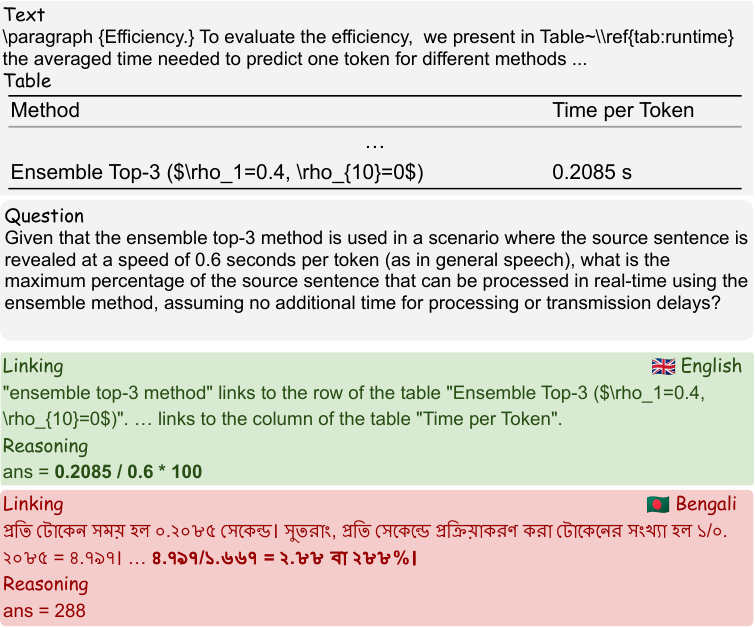}
    \caption{
    The case for the error type of "Formula".
    }
    \label{fig:case_formular}
\end{figure*}

\begin{figure*}[t]
    \centering
    \includegraphics[width=.85\linewidth]{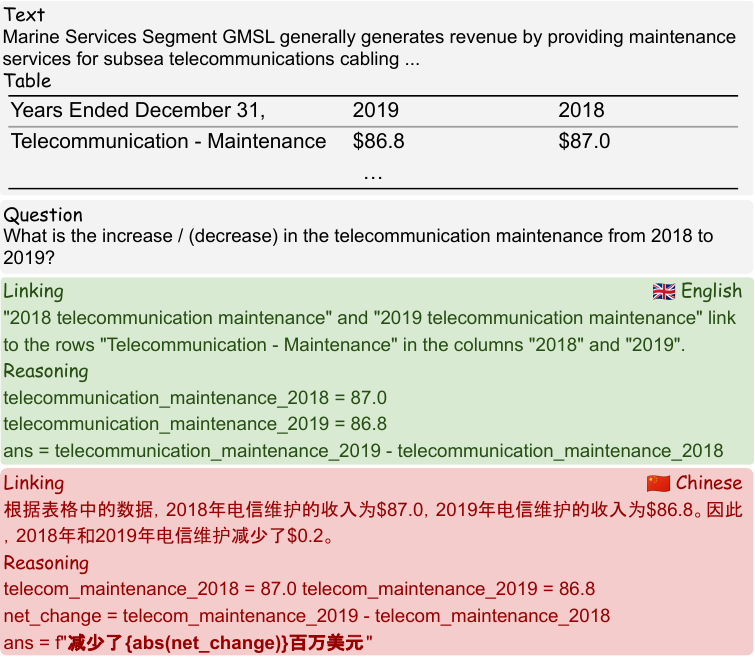}
    \caption{
    The case for the error type of "Redundancy".
    }
    \label{fig:case_redundancy}
\end{figure*}

In this subsection, we show the cases of error types corresponding to the analysis in \S\ref{subsec:Error Analysis}, as shown in Figure~\ref{fig:case_linking}, Figure~\ref{fig:case_formular}, and Figure~\ref{fig:case_redundancy}.

%% file: tab/comparison_tat.tex
\begin{tabular}{@{}llll@{}}
\toprule
\textbf{Dataset} & \textbf{Domain} & \textbf{Language}      \\ 
\midrule
GeoTSQA~\cite{li2021tsqa} & Geography & Chinese       \\
HybridQA~\cite{chen-etal-2020-hybridqa} & Wikipedia & English       \\
TAT-QA~\cite{zhu-etal-2021-tat} & Finance & English       \\
FinQA~\cite{chen-etal-2021-finqa} & Finance  & English       \\
QRData~\cite{liu-etal-2024-qrdata} & Cross & English       \\
DocMath-Eval~\cite{zhao-etal-2024-docmath}   & Finance  & English      \\
FinanceMATH~\cite{zhao-etal-2024-FinanceMATH} & Finance  & English      \\
SciTAT~\cite{zhang2024scitat} & Science  & English       \\ 
\midrule
\ourdataset  & Wikipedia + Finance + Science & Multilingual \\ 
\bottomrule
\end{tabular}

%% file: tab/prompt_baseline.tex
\begin{tabular}{p{0.9\textwidth}}
\toprule
\textbf{The prompt for Native-CoT} \\
\midrule
Lisez le texte et le tableau suivants, puis répondez à une question\\
Voici plusieurs exemples :\\
\\
---\\

\{Demonstrations\}

---\\

Sur la base des exemples ci-dessus, répondez à la question suivante.\\
Représentez votre réponse par : "Explication : <votre explication>\\
Réponse : <votre réponse>"\\
\\
\{Table\}\\
\{Paragraph\}\\
Question :\{Question\}\\
\bottomrule
\end{tabular}

\begin{tabular}{p{0.9\textwidth}}
\toprule
\textbf{The prompt for En-CoT} \\
\midrule
Read the following text and table, and then answer a question.\\
Here are several examples:\\
\\
---\\

\{Demonstrations\}

---\\
Based on the examples above, answer the following question.\\
Represent your answer with: "Explanation: <your explanation>\\
Answer: <your answer>"\\
\\
\{Table\}\\
\{Paragraph\}\\
Question :\{Question\}\\
\bottomrule
\end{tabular}

\begin{tabular}{p{0.9\textwidth}}
\toprule
\textbf{The prompt for Native-PoT} \\
\midrule
Lisez le texte et le tableau suivants, puis écrivez un code Python pour répondre à une question\\
Voici plusieurs exemples :\\
\\
---\\

\{Demonstrations\}

---\\

Sur la base des exemples ci-dessus, répondez à la question suivante avec un code Python.\\
Représentez votre réponse par : "and = <votre réponse>"\\
\\
\{Table\}\\
\{Paragraph\}\\
Question :\{Question\}\\
\bottomrule
\end{tabular}

\begin{tabular}{p{0.9\textwidth}}
\toprule
\textbf{The prompt for En-PoT} \\
\midrule
Read the following text and table, and then write a python code to answer a question\\
Here are several examples:\\
\\
---\\

\{Demonstrations\}

---\\
Based on the examples above, answer the following question with a Python code.\\
Represent your answer with: "ans = <your answer>"
\\
\{Table\}\\
\{Paragraph\}\\
Question :\{Question\}\\
\bottomrule
\end{tabular}

%% file: tab/prompt_our_method.tex
\begin{tabular}{p{0.9\textwidth}}
\toprule
\textbf{The prompt for \ourmethod} \\
\midrule
Please think in English and locate the relevant information from the text and table according to the question.\\
Here are several examples:\\
---\\
7. Nombre et coûts des employés...\\
| —                      | 2019   | 2018   |\\
| ---------------------- | ------ | ------ |\\
| —                      | Nombre | Nombre |\\
...\\
Question: Quelles sont les catégories d'employés listées dans le tableau ?\\
"Catégories des employés" links to the rows of the table "Opérations clients", "Produit et technologie", "Corporate" and the columns of the table "2019", "2018".\\
---\\
Le tableau suivant présente la répartition des revenus par catégorie et segment. ...\\
| Année se terminant le 31 décembre,    |          |         |\\
| ------------------------------------- | -------- | ------- |\\
|                                       | 2019     | 2018    |\\
...\\
Question: En 2019, combien de régions géographiques ont des revenus totaux supérieurs à 20 000 milliers de dollars\\
"2019" links to the column of the table "2019". "total revenues of geographic regions" links to the rows of the table "Total des revenus de l'Asie-Pacifique", "Total des revenus en Europe", "Total des revenus en Amérique du Nord".\\
---\\
Taux d'imposition effectif...\\
| —                                          | 31 décembre 2019 | 31 décembre 2018 |\\
...\\
Question: Quel a été le pourcentage de variation des pertes avant impôts en 2019 ?\\
"pérdidas antes de impuestos de 2019" y "pérdidas antes de impuestos de 2018" se vinculan a la parte del texto "In 2019 and 2018 we had pre-tax losses of \$19,573 and \$25,403, respectively".\\
---\\
Based on the examples above, analyze the question.\\
Please note that you **only** need to locate the relevant information, without performing additional calculations.\\
\{Table\}\\
\{Paragraph\}\\
Question :\{Question\}\\
\\
According to the relevant information, you should also think in English and write a python code to answer the question.\\
Here are several examples:\\
---\\
...\\
```python\\
ans = ['Opérations clients', 'Produit et technologie', 'Corporate']\\
```\\
---\\
...\\
```python\\
total\_revenues\_in\_all\_regions = \{'Asie-Pacifique': 6490, 'Europe': 36898, 'Amérique du Nord': 68024\}\\
regions\_have\_more\_than\_20000\_thousand\_total\_revenues = [k for k, v in total\_revenues\_in\_all\_regions.items() if v > 20000]\\
ans = len(regions\_have\_more\_than\_20000\_thousand\_total\_revenues)\\
```\\
---\\
...\\
```python\\
pre\_tax\_losses\_2018 = 25403  pre\_tax\_losses\_2019 = 19573\\
net\_change = pre\_tax\_losses\_2019 - pre\_tax\_losses\_2018\\
ans = net\_change / pre\_tax\_losses\_2018 * 100\\
```\\
---\\
Based on the examples above, answer the question with a Python code.\\
Please note:\\
1. In addition to numbers, try to use fr as the answer.\\
2. Keep your answer **short** with fewer statements.\\
3. Note the possible minus sign.\\
4. You MUST generate a Python code instead of returning the answer directly.\\
Represent your answer with: "ans = <your answer>"\\
\{Table\}\\
\{Paragraph\}\\
Question :\{Question\}\\
\bottomrule
\end{tabular}

%% file: tab/other_baselines.tex
\begin{tabular}{l|l|cccccc}
\toprule
\textbf{Model} & \textbf{Method} & \textbf{bn} & \textbf{de} & \textbf{en} & \textbf{es} & \textbf{fr} & \textbf{ja}\\
\midrule
\multirow{4}{*}{Llama3.1-8b} & Direct & $10.4/14.0$ & $12.8/17.7$ & $14.8/21.6$ & $13.6/21.1$ & $11.6/17.3$ & $10.4/12.3$ \\
 & Trans-CoT & $2.0/2.4$ & $15.2/16.0$ & $20.8/23.7$ & $18.8/20.6$ & $13.2/13.5$ & $9.6/11.1$  \\
 & Trans-PoT & $2.4/2.5$ & $20.4/21.2$ & $23.2/24.4$ & $21.2/21.6$ & $18.4/18.8$ & $10.8/11.1$ \\
 & \ourmethod & \bm{$20.0/21.3$} & \bm{$22.4/24.2$} & \bm{$27.6/31.9$} & \bm{$25.6/27.8$} & \bm{$20.0/22.4$} & \bm{$25.6/26.1$}  \\
\midrule
\multirow{4}{*}{Llama3.1-70b} & Direct & $12.4/17.4$ & $21.2/24.5$ & $22.0/26.6$ & $21.6/27.4$ & $18.0/22.3$ & $21.6/24.2$ \\
 & Trans-CoT & $4.4/4.9$ & $20.4/22.0$ & $25.6/29.3$ & $25.6/29.0$ & $16.4/18.1$ & $14.0/14.7$ \\
 & Trans-PoT & $3.2/3.4$ & $22.8/23.8$ & $30.4/32.9$ & $28.4/25.8$ & $22.8/23.7$ & $14.4/14.7$ \\
 & \ourmethod & \bm{$24.0/26.3$} & \bm{$28.0/31.3$} & \bm{$31.2/35.3$} & \bm{$29.4/34.6$} & \bm{$26.8/31.1$} & \bm{$26.8/29.4$} \\
\bottomrule
\end{tabular}

\begin{tabular}{l|l|cccccc}
\toprule
\textbf{Model} & \textbf{Method} & \textbf{ru} & \textbf{sw} & \textbf{te} & \textbf{th} & \textbf{zh} & \textbf{Avg.} \\
\midrule
\multirow{4}{*}{Llama3.1-8b} & Direct & $10.8/14.5$ & $9.6/14.7$ & $10.0/13.7$ & $12.0/14.1$ & $11.2/19.3$ & $11.6/16.4$ \\
 & Trans-CoT & $16.0/18.0$ & $9.2/10.2$ & $9.2/9.6$ & $11.6/13.1$ & $4.8/8.4$ & $11.9/13.3$ \\
 & Trans-PoT & $21.2/22.5$ & $16.0/16.4$ & $14.8/15.1$ & $13.6/14.9$ & $6.4/7.2$ & $15.3/15.8$ \\
 & \ourmethod & \bm{$25.2/27.0$} & \bm{$17.2/20.0$} & \bm{$14.4/15.2$} & \bm{$22.8/24.6$} & \bm{$23.6/28.0$} & \bm{$22.2/24.4$} \\
\midrule
\multirow{4}{*}{Llama3.1-70b} & Direct & $20.4/23.4$ & $20.0/23.3$ & $16.8/20.1$ & $20.4/23.5$ & $19.6/28.5$ & $19.5/23.7$ \\
 & Trans-CoT & $21.2/22.9$ & $17.6/19.6$ & $14.8/16.4$ & $19.6/21.9$ & $9.2/12.9$ & $17.0/18.4$ \\
 & Trans-PoT & $24.0/24.8$ & $20.0/20.9$ & $19.6/20.5$ & $18.0/19.5$ & $9.6/12.4$ & $19.6/20.4$ \\
 & \ourmethod & \bm{$28.8/33.5$} & \bm{$30.8/34.7$} & \bm{$22.8/25.9$} & \bm{$26.8/30.5$} & \bm{$28.0/34.9$} & \bm{$27.6/31.6$} \\
\bottomrule
\end{tabular}

%% file: fig/answer_source_8b.tex
\resizebox{0.75\linewidth}{!}{
\begin{tikzpicture}
\begin{polaraxis}[
    title={\empty},
    xlabel={\empty},
    ylabel={\empty},
    xtick={0,32.73,65.46,98.19,130.92,163.65,196.38,229.11,261.84,294.57,327.30},
    xticklabels={bn, de, en, es, fr, ja, ru, sw, te, th, zh},
    ytick={10,20,30,40},
    yticklabels={10,20,30,40},
    grid=both,
    tick label style={font=\small},
    major grid style={solid, gray}, 
    minor grid style={solid, gray}, 
    axis line style={draw=none},
    axis on top=true,                
    grid style={gray},              
    line join=bevel,
    tick align=outside,
]

\addplot[
    color=data_blue!300,
    mark=diamond,
    style={line width=1pt},
    fill=data_blue!250,fill opacity=0.2
] coordinates {
    (0,19.8)(32.73,20.9)(65.46,25.3)(98.19,22.0)(130.92,14.3)(163.65,23.1)(196.38,19.8)(229.11,18.7)(261.84,7.7)(294.57,22.0)(327.30,23.1)(0,19.8)
} -- cycle;

\addplot[
    color=reasoner_green!300,
    mark=square,
    style={line width=1pt},
    fill=reasoner_green!250,fill opacity=0.2
] coordinates {
    (0,11.1)(32.73,14.8)(65.46,25.9)(98.19,25.9)(130.92,11.1)(163.65,16.7)(196.38,18.5)(229.11,11.1)(261.84,16.7)(294.57,13.0)(327.30,9.3)(0,11.1)
} -- cycle;

\addplot[
    color=annotator_pink!150,
    mark=triangle,
    style={line width=1pt},
    fill=annotator_pink,fill opacity=0.2
] coordinates {
    (0,26.0)(32.73,28.0)(65.46,31.0)(98.19,30.0)(130.92,29.0)(163.65,34.0)(196.38,35.0)(229.11,20.0)(261.84,20.0)(294.57,30.0)(327.30,32.0)(0,26.0)
} -- cycle;

\legend{Table, Text, Hybrid}

\end{polaraxis}
\end{tikzpicture}
}

%% file: fig/answer_source_4o.tex
\resizebox{0.75\linewidth}{!}{
\begin{tikzpicture}
\begin{polaraxis}[
    title={\empty},
    xlabel={\empty},
    ylabel={\empty},
    xtick={0,32.73,65.46,98.19,130.92,163.65,196.38,229.11,261.84,294.57,327.30},
    xticklabels={bn, de, en, es, fr, ja, ru, sw, te, th, zh},
    ytick={10,20,30,40,50},
    yticklabels={10,20,30,40,50},
    grid=both,
    tick label style={font=\small},
    major grid style={solid, gray}, 
    minor grid style={solid, gray}, 
    axis line style={draw=none},
    axis on top=true,                
    grid style={gray},              
    line join=bevel,
    tick align=outside,
]

\addplot[
    color=data_blue!300,
    mark=diamond,
    style={line width=1pt},
    fill=data_blue!250,fill opacity=0.2
] coordinates {
    (0,29.7)(32.73,29.7)(65.46,34.1)(98.19,28.6)(130.92,26.4)(163.65,33.0)(196.38,30.8)(229.11,27.5)(261.84,26.4)(294.57,30.8)(327.30,26.4)(0,29.7)
} -- cycle;

\addplot[
    color=reasoner_green!300,
    mark=square,
    style={line width=1pt},
    fill=reasoner_green!250,fill opacity=0.2
] coordinates {
    (0,22.2)(32.73,31.5)(65.46,27.8)(98.19,25.9)(130.92,24.1)(163.65,27.8)(196.38,27.8)(229.11,24.1)(261.84,24.1)(294.57,22.2)(327.30,24.1)(0,22.2)
} -- cycle;

\addplot[
    color=annotator_pink!150,
    mark=triangle,
    style={line width=1pt},
    fill=annotator_pink,fill opacity=0.2
] coordinates {
    (0,34.0)(32.73,35.0)(65.46,40.0)(98.19,39.0)(130.92,36.0)(163.65,32.0)(196.38,34.0)(229.11,37.0)(261.84,34.0)(294.57,34.0)(327.30,37.0)(0,34.0)
} -- cycle;

\legend{Table, Text, Hybrid}

\end{polaraxis}
\end{tikzpicture}
}

%% file: fig/answer_source_en_CoT.tex
\resizebox{0.75\linewidth}{!}{
\begin{tikzpicture}
\begin{polaraxis}[
    title={\empty},
    xlabel={\empty},
    ylabel={\empty},
    xtick={0,32.73,65.46,98.19,130.92,163.65,196.38,229.11,261.84,294.57,327.30},
    xticklabels={bn, de, en, es, fr, ja, ru, sw, te, th, zh},
    ytick={10,20,30,40},
    yticklabels={10,20,30,40},
    grid=both,
    tick label style={font=\small},
    major grid style={solid, gray}, 
    minor grid style={solid, gray}, 
    axis line style={draw=none},
    axis on top=true,                
    grid style={gray},              
    line join=bevel,
    tick align=outside,
]

\addplot[
    color=data_blue!300,
    mark=diamond,
    style={line width=1pt},
    fill=data_blue!250,fill opacity=0.2
] coordinates {
    (0,14.3)(32.73,20.9)(65.46,25.3)(98.19,20.9)(130.92,23.1)(163.65,22.0)(196.38,27.5)(229.11,24.2)(261.84,17.6)(294.57,25.3)(327.30,17.6)(0,14.3)
} -- cycle;

\addplot[
    color=reasoner_green!300,
    mark=square,
    style={line width=1pt},
    fill=reasoner_green!250,fill opacity=0.2
] coordinates {
    (0,18.5)(32.73,18.5)(65.46,20.4)(98.19,18.5)(130.92,24.1)(163.65,25.9)(196.38,20.4)(229.11,16.7)(261.84,18.5)(294.57,11.1)(327.30,24.1)(0,18.5)
} -- cycle;

\addplot[
    color=annotator_pink!150,
    mark=triangle,
    style={line width=1pt},
    fill=annotator_pink,fill opacity=0.2
] coordinates {
    (0,21.9)(32.73,19.0)(65.46,23.8)(98.19,28.6)(130.92,26.7)(163.65,20.0)(196.38,25.7)(229.11,26.7)(261.84,21.9)(294.57,25.7)(327.30,24.8)(0,21.9)
} -- cycle;

\legend{Table, Text, Hybrid}

\end{polaraxis}
\end{tikzpicture}
}

%% file: fig/answer_source_en_PoT.tex
\resizebox{0.75\linewidth}{!}{
\begin{tikzpicture}
\begin{polaraxis}[
    title={\empty},
    xlabel={\empty},
    ylabel={\empty},
    xtick={0,32.73,65.46,98.19,130.92,163.65,196.38,229.11,261.84,294.57,327.30},
    xticklabels={bn, de, en, es, fr, ja, ru, sw, te, th, zh},
    ytick={10,20,30,40},
    yticklabels={10,20,30,40},
    grid=both,
    tick label style={font=\small},
    major grid style={solid, gray}, 
    minor grid style={solid, gray}, 
    axis line style={draw=none},
    axis on top=true,                
    grid style={gray},              
    line join=bevel,
    tick align=outside,
]

\addplot[
    color=data_blue!300,
    mark=diamond,
    style={line width=1pt},
    fill=data_blue!250,fill opacity=0.2
] coordinates {
    (0,22.0)(32.73,24.2)(65.46,30.8)(98.19,29.7)(130.92,25.3)(163.65,27.5)(196.38,28.6)(229.11,23.1)(261.84,24.2)(294.57,28.6)(327.30,22.0)(0,23.1)
} -- cycle;

\addplot[
    color=reasoner_green!300,
    mark=square,
    style={line width=1pt},
    fill=reasoner_green!250,fill opacity=0.2
] coordinates {
    (0,20.4)(32.73,18.5)(65.46,24.1)(98.19,18.5)(130.92,22.2)(163.65,22.2)(196.38,24.1)(229.11,22.2)(261.84,11.1)(294.57,14.8)(327.30,16.7)(0,20.4)
} -- cycle;

\addplot[
    color=annotator_pink!150,
    mark=triangle,
    style={line width=1pt},
    fill=annotator_pink,fill opacity=0.2
] coordinates {
    (0,26.7)(32.73,31.4)(65.46,34.3)(98.19,30.5)(130.92,32.4)(163.65,31.4)(196.38,33.3)(229.11,31.4)(261.84,29.5)(294.57,27.6)(327.30,36.2)(0,28.6)
} -- cycle;

\legend{Table, Text, Hybrid}

\end{polaraxis}
\end{tikzpicture}
}

%% file: fig/answer_type_8b.tex
\resizebox{0.75\linewidth}{!}{
\begin{tikzpicture}
\begin{polaraxis}[
    title={\empty},
    xlabel={\empty},
    ylabel={\empty},
    xtick={0,32.73,65.46,98.19,130.92,163.65,196.38,229.11,261.84,294.57,327.30},
    xticklabels={bn, de, en, es, fr, ja, ru, sw, te, th, zh},
    ytick={20,40,60,80,100},
    yticklabels={20,40,60,80,100},
    grid=both,
    tick label style={font=\small},
    major grid style={solid, gray}, 
    minor grid style={solid, gray}, 
    axis line style={draw=none},
    axis on top=true,                
    grid style={gray},              
    line join=bevel,
    tick align=outside,
]

\addplot[
    color=data_blue!300,
    mark=diamond,
    style={line width=1pt},
    fill=data_blue!250,fill opacity=0.2
] coordinates {
    (0,8.8)(32.73,10.9)(65.46,16.3)(98.19,12.9)(130.92,8.8)(163.65,10.2)(196.38,11.6)(229.11,8.8)(261.84,6.1)(294.57,9.5)(327.30,8.8)(0,8.8)
} -- cycle;

\addplot[
    color=reasoner_green!300,
    mark=square,
    style={line width=1pt},
    fill=reasoner_green!250,fill opacity=0.2
] coordinates {
    (0,41.0)(32.73,39.8)(65.46,43.4)(98.19,42.2)(130.92,37.3)(163.65,50.6)(196.38,48.2)(229.11,30.1)(261.84,28.9)(294.57,43.4)(327.30,47.0)(0,41.0)
} -- cycle;

\addplot[
    color=annotator_pink!150,
    mark=triangle,
    style={line width=1pt},
    fill=annotator_pink,fill opacity=0.2
] coordinates {
    (0,30.0)(32.73,70.0)(65.46,90.0)(98.19,100)(130.92,60.0)(163.65,70.0)(196.38,60.0)(229.11,50.0)(261.84,30.0)(294.57,70.0)(327.30,70.0)(0,30.0)
} -- cycle;

\legend{Span, Arithmetic, Count}

\end{polaraxis}
\end{tikzpicture}
}

%% file: fig/answer_type_4o.tex
\resizebox{0.75\linewidth}{!}{
\begin{tikzpicture}
\begin{polaraxis}[
    title={\empty},
    xlabel={\empty},
    ylabel={\empty},
    xtick={0,32.73,65.46,98.19,130.92,163.65,196.38,229.11,261.84,294.57,327.30},
    xticklabels={bn, de, en, es, fr, ja, ru, sw, te, th, zh},
    ytick={20,40,60,80,100},
    yticklabels={20,40,60,80,100},
    grid=both,
    tick label style={font=\small},
    major grid style={solid, gray}, 
    minor grid style={solid, gray}, 
    axis line style={draw=none},
    axis on top=true,                
    grid style={gray},              
    line join=bevel,
    tick align=outside,
]

\addplot[
    color=data_blue!300,
    mark=diamond,
    style={line width=1pt},
    fill=data_blue!250,fill opacity=0.2
] coordinates {
    (0,17.0)(32.73,16.3)(65.46,21.0)(98.19,17.0)(130.92,12.9)(163.65,21.1)(196.38,15.6)(229.11,18.4)(261.84,15.6)(294.57,15.0)(327.30,15.6)(0,17.0)
} -- cycle;

\addplot[
    color=reasoner_green!300,
    mark=square,
    style={line width=1pt},
    fill=reasoner_green!250,fill opacity=0.2
] coordinates {
    (0,51.8)(32.73,57.8)(65.46,55.4)(98.19,56.6)(130.92,54.2)(163.65,48.2)(196.38,55.4)(229.11,49.4)(261.84,50.6)(294.57,56.6)(327.30,53.0)(0,51.8)
} -- cycle;

\addplot[
    color=annotator_pink!150,
    mark=triangle,
    style={line width=1pt},
    fill=annotator_pink,fill opacity=0.2
] coordinates {
    (0,70.0)(32.73,90.0)(65.46,90.0)(98.19,90.0)(130.92,100)(163.65,70.0)(196.38,90.0)(229.11,90.0)(261.84,70.0)(294.57,70.0)(327.30,90.0)(0,70.0)
} -- cycle;

\legend{Span, Arithmetic, Count}

\end{polaraxis}
\end{tikzpicture}
}

%% file: fig/answer_type_en_CoT.tex
\resizebox{0.75\linewidth}{!}{
\begin{tikzpicture}
\begin{polaraxis}[
    title={\empty},
    xlabel={\empty},
    ylabel={\empty},
    xtick={0,32.73,65.46,98.19,130.92,163.65,196.38,229.11,261.84,294.57,327.30},
    xticklabels={bn, de, en, es, fr, ja, ru, sw, te, th, zh},
    ytick={20,40,60,80,100},
    yticklabels={20,40,60,80,100},
    grid=both,
    tick label style={font=\small},
    major grid style={solid, gray}, 
    minor grid style={solid, gray}, 
    axis line style={draw=none},
    axis on top=true,                
    grid style={gray},              
    line join=bevel,
    tick align=outside,
]

\addplot[
    color=data_blue!300,
    mark=diamond,
    style={line width=1pt},
    fill=data_blue!250,fill opacity=0.2
] coordinates {
    (0,6.4)(32.73,8.9)(65.46,12.7)(98.19,15.9)(130.92,13.4)(163.65,14.6)(196.38,13.4)(229.11,12.7)(261.84,10.8)(294.57,12.7)(327.30,12.1)(0,6.4)
} -- cycle;

\addplot[
    color=reasoner_green!300,
    mark=square,
    style={line width=1pt},
    fill=reasoner_green!250,fill opacity=0.2
] coordinates {
    (0,37.3)(32.73,32.5)(65.46,39.8)(98.19,31.3)(130.92,41.0)(163.65,31.3)(196.38,39.8)(229.11,37.3)(261.84,30.1)(294.57,34.9)(327.30,33.7)(0,37.3)
} -- cycle;

\addplot[
    color=annotator_pink!150,
    mark=triangle,
    style={line width=1pt},
    fill=annotator_pink,fill opacity=0.2
] coordinates {
    (0,50.0)(32.73,80.0)(65.46,60.0)(98.19,80.0)(130.92,70.0)(163.65,60.0)(196.38,90.0)(229.11,80.0)(261.84,70.0)(294.57,70.0)(327.30,80.0)(0,50.0)
} -- cycle;

\legend{Span, Arithmetic, Count}

\end{polaraxis}
\end{tikzpicture}
}

%% file: fig/answer_type_en_PoT.tex
\resizebox{0.75\linewidth}{!}{
\begin{tikzpicture}
\begin{polaraxis}[
    title={\empty},
    xlabel={\empty},
    ylabel={\empty},
    xtick={0,32.73,65.46,98.19,130.92,163.65,196.38,229.11,261.84,294.57,327.30},
    xticklabels={bn, de, en, es, fr, ja, ru, sw, te, th, zh},
    ytick={20,40,60,80,100},
    yticklabels={20,40,60,80,100},
    grid=both,
    tick label style={font=\small},
    major grid style={solid, gray}, 
    minor grid style={solid, gray}, 
    axis line style={draw=none},
    axis on top=true,                
    grid style={gray},              
    line join=bevel,
    tick align=outside,
]

\addplot[
    color=data_blue!300,
    mark=diamond,
    style={line width=1pt},
    fill=data_blue!250,fill opacity=0.2
] coordinates {
    (0,7.6)(32.73,8.9)(65.46,12.7)(98.19,11.5)(130.92,10.8)(163.65,11.5)(196.38,12.1)(229.11,7.6)(261.84,10.2)(294.57,10.2)(327.30,10.2)(0,8.3)
} -- cycle;

\addplot[
    color=reasoner_green!300,
    mark=square,
    style={line width=1pt},
    fill=reasoner_green!250,fill opacity=0.2
] coordinates {
    (0,48.2)(32.73,50.6)(65.46,57.8)(98.19,49.4)(130.92,50.6)(163.65,53.0)(196.38,55.4)(229.11,53.0)(261.84,44.6)(294.57,48.2)(327.30,51.8)(0,50.6)
} -- cycle;

\addplot[
    color=annotator_pink!150,
    mark=triangle,
    style={line width=1pt},
    fill=annotator_pink,fill opacity=0.2
] coordinates {
    (0,70.0)(32.73,90.0)(65.46,90.0)(98.19,100.0)(130.92,100.0)(163.65,90.0)(196.38,100.0)(229.11,100.0)(261.84,60.0)(294.57,70.0)(327.30,80.0)(0,70.0)
} -- cycle;

\legend{Span, Arithmetic, Count}

\end{polaraxis}
\end{tikzpicture}
}